\documentclass[12pt]{article}

\usepackage{newtxtext,newtxmath}

\usepackage{graphicx}

\usepackage[letterpaper,margin=1in]{geometry}

\renewenvironment{abstract}
    {\quotation}
    {\endquotation}

\date{}

\makeatletter
\renewcommand{\fnum@figure}{\textbf{Figure \thefigure}}
\renewcommand{\fnum@table}{\textbf{Table \thetable}}
\makeatother

\usepackage{scicite}

\usepackage{url}

\usepackage[utf8]{inputenc}
\usepackage[T1]{fontenc}
\usepackage[english]{babel}

\usepackage{graphicx}  
\usepackage{epstopdf} 

\usepackage[nointegrals]{wasysym} 
\usepackage{textcomp}  

\usepackage{amsfonts}

\usepackage{amsmath}
 \usepackage[normalem]{ulem}

\usepackage{amssymb}
\usepackage{mathtools}
\usepackage{amsthm}
\usepackage{mathrsfs}
\usepackage{dsfont}
\usepackage{bbm}
\usepackage{nicefrac}
\usepackage{xfrac}

\usepackage{graphicx}
\usepackage{float}
\usepackage{subcaption}
\usepackage{caption}

\usepackage{booktabs}
\usepackage{threeparttable}
\let\oldtnote\tnote  
\usepackage{array, multirow, multicol}
\let\tnote\oldtnote  

\usepackage{listings}

\let\tnote\relax
\usepackage{xcolor}
\usepackage{wasysym}

\usepackage{balance}
\usepackage{flushend}
\usepackage{titlesec}
\usepackage[none]{hyphenat}
\usepackage[final]{pdfpages}

\usepackage{cite}
\let\tnote\relax
\usepackage{hyperref}
\hypersetup{colorlinks=true,urlcolor=blue,citecolor=red}

\usepackage[ruled, linesnumbered]{algorithm2e}

\usepackage{lineno}
\usepackage{etoolbox}
\usepackage{soul}
\usepackage{resizegather}
\usepackage{eso-pic}

\let\amssymbboxplus\boxplus
\let\amssymbboxminus\boxminus

\renewcommand{\boxplus}{\mathbin{\mathop\amssymbboxplus}}
\renewcommand{\boxminus}{\mathbin{\mathop\amssymbboxminus}}

\DeclareMathAlphabet{\mathmybb}{U}{bbold}{m}{n}

\usepackage[dvipsnames]{xcolor}
\usepackage{titlesec}           

\titleformat{\section}
  {\normalfont\Large\bfseries\MakeUppercase} 
  {\thesection}                                       
  {1em}                                               
  {}                                                  

\titleformat{\subsection}
  {\normalfont\large\bfseries}
  {\thesubsection}
  {1em}
  {}

\titleformat{\subsubsection}
  {\normalfont\normalsize\bfseries\itshape}
  {\thesubsubsection}
  {1em}
  {}

\usepackage{newfloat}
\DeclareFloatingEnvironment[name={Movie},fileext={lov}]{movie}
\usepackage{xurl}

\def\scititle{MilliWatt Ultrasound for Navigation in Visually Degraded Environments on Palm-Sized Aerial Robots}
\title{\bfseries \boldmath \scititle}

\author{
    Manoj~Velmurugan$^{1}$,
    Phillip~Brush$^{1}$,
    Colin~Balfour$^{1}$,\and
        Richard~J.~Przybyla$^{2}$,
        Nitin~J.~Sanket$^{1\ast}$\and
    \small$^{1}$Perception and Autonomous Robotics (PeAR) Group, \\ \small Worcester Polytechnic Institute, Worcester, MA 01609, USA. \\
    \small$^{2}$TDK InvenSense, Berkeley, CA 94710, USA.\\
    \small$^\ast$Corresponding author. Email: \texttt{nitin@wpi.edu}
    }

\newcommand{\textred}[1]{#1}

\begin{document} 

\AddToShipoutPictureFG*{%
  \AtPageUpperLeft{%
    \hspace*{1in}\raisebox{-1in}{%
      \parbox{1.1\textwidth}{\small
      PUBLISHED IN SCIENCE ROBOTICS, 2026. DOI: \url{https://doi.org/10.1126/scirobotics.adz9609}}
    }%
  }%
}

\maketitle

\begin{abstract} \bfseries \boldmath
Tiny palm-sized aerial robots possess exceptional agility and cost-effectiveness in navigating confined and cluttered environments. However, their limited payload capacity directly constrains the sensing suite on-board the robot, thereby limiting critical navigational tasks in \textred{Global Positioning System (GPS)-denied} wild scenes. Common methods for obstacle avoidance use cameras and \textred{LIght Detection And Ranging (LIDAR)}, which become ineffective in visually degraded conditions such as low visibility, dust, fog or darkness. Other sensors, such as \textred{RAdio Detection And Ranging (RADAR)}, have high power consumption, making them unsuitable for tiny aerial robots. Inspired by bats, we propose Saranga, a low-power ultrasound-based perception stack that localizes obstacles using a dual sonar array. We present two key solutions to combat the low \textcolor{black}{Peak Signal-to-Noise Ratio of $-4.9$~{decibels}}: physical noise reduction and a deep learning based denoising method.  Firstly, we present a practical way to block propeller induced ultrasound noise on the weak echoes. The second solution is to train a neural network to utilize the \textcolor{black}{long horizon of ultrasound echoes} for finding signal patterns under high amounts of uncorrelated noise where classical methods were insufficient. We generalize to the \textcolor{black}{real world by using a synthetic data generation pipeline and limited real noise data for training}. We enable a palm-sized aerial robot to navigate in visually degraded conditions of \textcolor{black}{dense fog}, darkness, and snow in a cluttered environment with thin and transparent obstacles using only on-board sensing and computation. We provide extensive \textcolor{black}{real world} results to demonstrate the efficacy of our approach.  
\end{abstract}




\section*{INTRODUCTION}
The bumblebee bat, weighing only 2~g can navigate in deep, dark caves under dusty conditions with remarkable accuracy by perceiving something as small as 8~mm using \textcolor{black}{echolocation}\cite{griffin_sensitivity_1958}. With a simple bi-sonar sensory apparatus and a mere 2~million neurons, these expert fliers thrive in conditions most large animals cannot through \textcolor{black}{parsimony (utilizing minimal sensory and computational resources) by using ingenious solutions} to complex navigational problems. Most living agents possess remarkable visual acuity centered around the electromagnetic wave spectrum (visible, infrared, or ultraviolet light). In contrast, bats utilize ultrasound-based echolocation (sending short chirps and listening to echoes) and achieve similar or better accuracies in visually degraded environments such as dark and dusty caves. Autonomous aerial robots today utilize a myriad of sensors tailored for each sensing scenario, ranging from LIDARs\cite{ren_science_2025}, RADARs\cite{guido_radar}, tactile sensors\cite{hamaza_2025} and infrared-based depth cameras\cite{gao_sci}, to traditional cameras\cite{Raju2024EdgeFlowNet} or even event cameras\cite{falanga_science_2020} to achieve autonomy with agility and accuracy. However, despite major advances in technology \cite{de_croon_2022}, today's autonomous aerial robots are rarely seen deployed in highly cluttered environments under harsh environmental conditions such as fog, dust, smoke, low-light and/or snow, limiting widespread usage for \textred{saving lives} in search and rescue where these scenarios are commonly encountered\cite{kailin2024, Shizhe2019, Alexis_Darpa2022}. \textcolor{black}{We attribute this limited deployment to four key factors:}  \textcolor{black}{robot size}, \textcolor{black}{robot cost}, \textcolor{black}{operational sensing power requirements}, and safety considerations. \textcolor{black}{Currently available commercial autonomous drones such as the Skydio X10 \cite{skydioX10}, DJI Matrice 4TD \cite{djiMatrice4TD}, and EVO Max Series V2 \cite{evoMaxV2} have diagonal wheelbases ranging from $0.467$~m to $0.8$~m, weigh between $1.7$~kg and $2.2$~kg, cost between 5{,}500~USD and 15{,}000~USD, utilize upwards of 20~W of sensing power, and are unsafe around densely cluttered environments due to large propellers.}\\
\textred{In this article}, we present a system on a tiny autonomous quadrotor (measuring \textcolor{black}{0.16~m} in size, costing only 400~USD and utilizing a mere 1.2~mW of sensing power) equipped with a dual low-power ultrasound sensor suite to navigate in harsh conditions of fog, low-light, snow, and \textcolor{black}{darkness} using only on-board sensing and computation with no reliance on any external infrastructure. The choice of ultrasound sensors is centered around parsimony and ultrasound's properties of working under a myriad of conditions. We tackle challenges of an extremely low \textcolor{black}{\textred{Peak Signal-to-Noise Ratio (PSNR)} ($-4.9$~dB)} that have rendered ultrasound ineffectual for widespread aerial autonomy until now. Drawing inspiration from bats, we utilize the data contained in Interaural Time Disparity (ITD) \cite{borina_neural_2011} to localize and dodge obstacles for navigation in the wild. We propose two key ideas in this work: \textred{A deep neural network called Saranga that enables denoising weak echo signals and a physical noise reduction by blocking propeller-generated ultrasound noise.} With these two methods, we enable the localization of obstacles with weak echoes for navigation in various visually degraded conditions in the wild. Our work (and the network) called Saranga is named after the celestial bow of Vishnu from Hindu Mythology, which possesses the ability to penetrate anything.  In our work, ultrasound is used to ``penetrate'' through visually challenging conditions such as \textcolor{black}{dense fog}, dust, \textred{darkness} and snow. Furthermore, using only 1.2~mW of sensing power to enable operation, our system showcases exceptional efficiency.


\subsection*{Related work}

Obstacle avoidance on aerial robots traditionally uses sensors such as cameras, event cameras, RADAR and/or LIDAR \cite{Chandran2023ReviewOT, xu_obs_avoid, falanga_science_2020, ren_science_2025, ajna_23, GREEFF20209412, Raju2024EdgeFlowNet, sanket_2020, guido_radar, floreano2015science, dharmadhikari2021}. Although these sensors are accurate under normal conditions, they fail in one/more of the common visually degraded scenarios, including direct light, darkness, \textcolor{black}{snow}, glass, \textcolor{black}{fog and/or dust} (See Figs.~\ref{fig:sensor_comparison_complete}, \ref{fig:indoor_nav_image}). \textcolor{black}{\textred{Although} RADAR systems perform well in many scenarios, they often struggle to detect objects made of dielectric materials such as plastics \cite{infineon-technologies-ag-2020, Bur1985}, as is common in various building materials.} Furthermore, LIDARs and RADARs consume tens of Watts of power for sensing. Operations in these visually degraded scenarios are critical for many missions, such as search and rescue and cave exploration, which are currently tackled by using a multitude of sensors on the robot \cite{Alexis_Darpa2022, kostas_thermal2020, kostas2017, saska_2021, Almalioglu2022, papachristos2018}. 

In general, acoustic sensors have been studied by the aerial robotics community for obstacle avoidance \cite{muller_batdeck_2024, muller_superbat, muller_safe_2014, dumbgen_2023, ahn_2024}, altitude estimation\cite{calkins_distance_2021}, and odometry\cite{muller_batdeck_ultra}. These acoustic sensing approaches can be broadly classified into two categories: auditory range sensing (human) and ultrasound sensing. Blind as a Bat \cite{dumbgen_2023} utilized a buzzer and microphones on a Crazyflie and achieved centimeter-level wall localization. However, this has been proven to work indoors for large obstacles like walls, and only for \textred{nano} aerial robots with brushed motors \textred{and} low noise levels. These tiny robots currently cannot perform complex navigational tasks. Additionally, the system is \textred{blaring} and disruptive to people/animals around. Since propellers also generate substantial noise, Calkins et al. \cite{calkins_distance_2021} utilized a pair of microphones to estimate the distance to the ground (up to 140~mm) solely using propeller noise as an excitation mechanism.

\textcolor{black}{Safe and Sound \cite{muller_safe_2014} used four orthogonal ultrasonic sensors to maintain position within an empty room of four walls on a quadrotor. The obstacle avoidance in this approach was simply to maintain a set distance from each wall. Furthermore, no navigation performance was demonstrated in a complex scene or with varying ecological factors. }

BatDeck \cite{muller_batdeck_2024} demonstrated robust ultrasound-based avoidance using a single front-facing sensor indoors. Since the obstacle direction information was unavailable, the robot avoided obstacles by turning randomly. Such a method results in suboptimal trajectories that waste precious battery power and flight time. Additionally, its algorithm can cause the quadrotor to almost completely stop during navigation while making turns. SuperBat \cite{muller_superbat} overcomes some of these limitations by adding a laser Time of Flight imaging rangefinder; however, this adds more complexity without addressing the actual challenges in visually degraded scenarios like fog and glass. 

Ultrasound-based obstacle avoidance is common in mobile (generally ground) robots and is implemented through different modalities: \textred{trilateration \cite{walter_locating_2014, nagashima_1992, Kreczmer10, THONGUN2015480}, array sensors \cite{jimenez_2005, strakowski2006, dias_2020, steckel_broadband_2013, jimenez_2015, microsoft_2014, laurijssen_2024, marzo_2018, steckel_2024}, multiple narrow beam rangefinders \cite{morovec_1985, jungdong_mobile_2007, brutus_2000}, and deep learning-based scene reconstruction \cite{catChatter, batVision}.} Utilizing low-power ultrasound sensors on aerial robots poses a peculiar problem of propeller-induced noise. For simple obstacle detection methods, such as detect and stop, they can work on tiny aerial robots with simple filtering methods \cite{muller_batdeck_2024}. However, these robots cannot perform functional navigational tasks. Furthermore, palm-sized aerial robots of sizes up to 100~mm in size and weights up to 500~g have \textred{substantial} noise levels, most of the time enough to ``drown'' out weak signals from echoes over a meter away \textred{(See Noise reduction section)}.

\textcolor{black}{
        The effect of noise on acoustic sensors mounted on aerial robots remains an understudied area for ultrasonic echolocation systems. \textred{Although} several works exist in the audible acoustic range targeting applications such as audio denoising \cite{azarang2020, xu2020listening} and sound source localization \cite{chevtchenko_2025, nakul2021_owlet}, these methods typically require an array of eight or more microphones \cite{khan_2025} and focus primarily on direction-of-arrival estimation using methods like delay-sum beamforming \cite{mars_2018} and \textred{Multiple Signal Classification (MUSIC)} \cite{barabell_1984}. However, echolocation for navigation on aerial platforms faces fundamental challenges from propeller-induced noise \cite{deleforge_2020}, as the signal power drops by a factor of four compared to source localization due to the doubled path length and additional losses from reflecting surface properties.\\
Ears in the Sky \cite{michez_2021} analyzed how noise levels decrease with distance and proposed suspending microphones several meters below the aerial robot, an approach infeasible for tiny robots. Huang et al. \cite{huang_2020} developed a sound-based local positioning system and extensively analyzed propeller-induced noise across varying thrust levels, propeller sizes, frequency bands, and microphone locations. They mounted microphones 30~cm from the robot's center to reduce \textred{the influence of noise}, with ultrasound signals transmitted from an external source without power or weight constraints. \textred{Although} these approaches suit large aerial robots, physically separating sensors from noise sources is impractical for palm-sized robots that must navigate cluttered environments. 
}

It is important to note that there is a wealth of work in the medical imaging literature on ultrasound, which we do not discuss here. The medical imaging works are centered on contact-based imaging, where wave-medium interactions are very different from the interaction of ultrasound in air, like the one used in this work.


\subsection*{Problem definition and contributions}\label{sec:definition}
A quadrotor is present in an unknown scene with static obstacles of unknown shapes, sizes, materials and locations under harsh, visually degraded conditions with \textcolor{black}{dense fog}, darkness, dust and/or snow. The goal is to navigate towards a \textcolor{black}{given} direction while avoiding obstacles using only on-board sensing (dual low-power ultrasound chirp sensors) and computation. We answer the following questions: \textred{What practical considerations are required to utilize low-power ultrasound sensors effectively on quadrotors for obstacle avoidance? What representation do we need for the ultrasound signals to be able to detect obstacles?} A summary of our contributions is as follows: We propose a method to localize obstacles using a neural network-based solution for denoising, which we call Saranga, followed by practical performance improvements using a physical blocking structure to reduce the effect of propeller-induced noise. We perform extensive \textcolor{black}{real world} experiments demonstrating the efficacy of our approach in visually degraded conditions with \textcolor{black}{dense fog}, darkness, dust and \textcolor{black}{snow} (See Movie.~\ref{fig:banner} and Fig.\ref{fig:sensor_comparison_complete}).

\section*{RESULTS}

\subsection*{Quadrotor platform}

The aerial robot used in the experiments is a custom-built quadrotor platform called PeARBat160 (Movie.~\ref{fig:banner}). The platform used an X-shaped carbon fiber frame with a diagonal wheelbase of 160~mm. The drivetrain consisted of four HQProp 3$\times$4$\times$3" tri-blade propellers and T-Motor 1507 3800KV motors mated to a Hobbywing XRotor Micro 40A 6S BLHeli32 4-in-1 ESC (Electronic Speed Controller). We used a Matek Systems H743-Mini flight controller running ArduCopter v4.5.7 firmware. The robot is used in guided mode for autonomous operation.
\\
We utilized two TDK\texttrademark{} \textred{InvenSense\texttrademark{} ICU30201} \cite{przybyla2023mass} low-power ultrasound chirp sensors with a $140\times 57^\circ$ acoustic horn, which were synchronized using a Teensy 4.1 microcontroller. Sensor data was processed onboard using the Google Coral Mini\texttrademark{}\cite{coral_mini} development board running Mendel GNU/Linux 5 OS. The robot had an ARKFlow optical flow sensor (for low-cost 2D velocity estimates) and a bottom-facing ICU30201 sensor mated to a 45$^\circ$ acoustic horn for altitude measurements). All the autonomy software was run on \textred{Robot Operating System (ROS)} 2 Humble, with the data processing, perception, and planning algorithms run as individual ROS nodes. Navigation commands were computed at approximately $16~\mathrm{Hz}$ and transmitted to the flight controller over the MAVLink protocol \cite{mavlink}. The quadrotor had a five minute flight time and an All Up Weight (AUW) of 460~g with an 850~mAh 4S \textred{Lithium Polymer (LiPo)} battery. Note that all the obstacle avoidance was solely done using the front-facing ultrasound sensors using our Saranga pipeline (Fig.~\ref{fig:sensor_comparison_complete}\textcolor{red}{A}).

\textred{We performed two sets of experiments, indoor and outdoor, under various scenarios to test the efficacy of our approach under challenging conditions. These experiments, along with the metrics of success, are described in the following sections}. We performed exhaustive comparisons of our approach with other common robotics sensors for obstacle localization under various metrics of accuracy and operability.

\subsection*{Sensor, perception and navigation configuration details}

We empirically observed that a sensing range of 1.5 m was sufficient to dodge obstacles with a large enough time window for our fastest run (2 m/s). To this end, we set $N=512$ samples, which corresponds to a 1.66 m sensing range. The choice of $N=512$ spurs from the fact that power-of-two sizes are better suited for memory access and cache efficiency \cite{prgflow}.  

We stacked \(M = 32\) consecutive measurement cycles (referred to as \textred{history}), corresponding to  $\sim 0.82~\mathrm{s}$ of temporal context for the network. With $M=32$, the overall perception-planner stack runs at 16Hz. Increasing the history length to \(M = 64\) did not yield any noticeable improvement in obstacle avoidance performance, but reduced the effective perception–planner stack rate to about \(5\text{--}8~\mathrm{Hz}\).

\subsection*{Evaluation methodology}
We evaluated our perception and navigation algorithms on both indoor and outdoor settings. For each experiment, an obstacle course with obstacles of different shapes (cylindrical, cuboidal) and materials (transparent objects, natural trees, thin metal poles) was set up. Each obstacle course was evaluated for \textcolor{black}{a minimum of} 20 trials. \textcolor{black}{The number of trials was dictated by the number of flights that was achieved using six fully charged batteries. All the trials were run continuously per battery charge until depletion. When this resulted in more than 20 trials, all the runs are reported.} Navigating through the obstacle course without any collision is called a success. Hitting obstacles is marked as a failed run. Success rate is the ratio of the number of successful trials to the total trials. 
Traversability measures how densely obstacles are packed relative to the robot. Since the navigation policy does not control yaw, the robot’s heading is fixed to point straight ahead. In each trial, the robot was placed at a random location and moves forward. The image plane was divided into seven horizontal sectors (each spanning 20°), and the maximum traversable distance before encountering an obstacle was computed using the method from~\cite{guido_tv}. \textcolor{black}{The traversability score for each scene is computed by simulating 1100 robot trajectories and then averaging the traversable distance over samples collected as described in \cite{guido_tv}. We used a digital twin of the real scene based on COLMAP \cite{schoenberger2016sfm} for this computation. These samples are independent of the trials later discussed in the paper, which are used to evaluate the efficacy of Saranga.} Lower \textcolor{black}{traversability} scores indicate more cluttered, densely packed environments.

\textcolor{black}{The results presented in this section are summarized in Fig.~\ref{fig:success_rates}, along with additional environmental conditions such as brightness, humidity, temperature, and wind-speed that may affect the performance of sensors. We refer the readers to \textred{Supplementary Movies} \textcolor{red}{S1}, \textcolor{red}{S2} for a demonstration of our experiments.}

\subsection*{Indoor experiments}
All indoor experiments were conducted in the PeAR Washburn flying space at Worcester Polytechnic Institute, which has a netted flying space of $11~\mathrm{m} \times 4.5~\mathrm{m} \times 3.65~\mathrm{m}$. Ground truth pose estimates for comparison and plots are provided by a motion capture setup with a ring of 14 Vicon\texttrademark{}  Vero V2.2 cameras around the flying space. Note that we did not use the motion capture data for any of the autonomy algorithms, and all our results are based on onboard sensing and computation without any external infrastructure. 

\subsubsection*{Transparent obstacles}
Although LIDARs and other depth sensors work well across many common scenarios, they often struggle in visually degraded environments and objects transparent to visible light, such as glass, water, and plastics such as acrylic. These materials have become increasingly common in building materials due to lighting and thermal efficiency \cite{Bernd_2017} and the possibility of harnessing solar energy. Although vision-centric datasets have been proposed to tackle these problems \cite{Bernd_2017}, they still remain challenging for robot navigation applications. \textcolor{black}{To demonstrate the ultrasound sensor's advantage in detecting transparent obstacles}, we constructed an environment (Fig.~\ref{fig:indoor_nav_image}\textred{A}) with \textred{transparent ``walls'' made of thin, transparent thermoplastic} (thickness $<0.02$~mm),  which remain undetectable even to $76$--$81$~GHz radar systems, as shown in Fig.~\ref{fig:sensor_comparison_complete}\textcolor{red}{C}. The transparent films were stretched taut to minimize flutter caused by the propeller downwash. The transparent wall dimensions were in the range of $\{100,160\}\times\{125,150\}\times 0.002$~cm (length $\times$ height $\times$ thickness). The obstacles were arranged such that the course was non-traversable without detecting and avoiding them; the orientations and locations were randomized in every trial. We observed a success rate of $77.27\%$ across $22$~runs.

\subsubsection*{Thin obstacles}
Thin objects are particularly challenging for detection and dodging \cite{ren_science_2025}. Specially, thin branches and small poles pose dangers for \textcolor{black}{real world} deployment of aerial robots as they are hard to detect but can cause \textred{catastrophic} damage to the robots.  Although methods based on LIDAR \cite{ren_science_2025} and cameras \cite{madaan_2017} have been proposed for detection of thin wires, they utilize large amounts of sensing and computation power. Particularly for low-power ultrasound sensors, the reflected echoes are particularly weaker since the reflection surface area is small. By combining our deep network, Saranga, with physical noise reduction solutions, we address these limitations. In this experiment, seven thin obstacles, made of \textred{either Poly Vinyl Chloride (PVC) or aluminum, } were spread out in the environment (Fig.~\ref{fig:indoor_nav_image}\textcolor{red}{B}). 
The obstacles dimensions were in the range of $\{2,6\}\times\{152,183\}\times\{1.5,6\}$~cm (length $\times$ height $\times$ thickness).
We achieved a success rate of $80.95\%$ in the $21$~trials.

\subsubsection*{Snow}
We constructed a scenario with five cuboidal obstacles that were \textred{randomly distributed throughout} the flying space (Fig.~\ref{fig:indoor_nav_image}\textcolor{red}{C}). The obstacles' dimensions are in the range of $\{46,67\}\times\{115,138\}\times\{19,25\}$~cm (length $\times$ height $\times$ thickness) and were placed with arbitrary orientations and locations, changed over trials. Every flight was initialized with a random position in the obstacle course, in such a way that any flight along a straight trajectory would result in a collision in case of missed detection. The aim of this experiment was to test our approach in an environment with falling \textcolor{black}{artificial snow (we will refer to this as snow throughout the paper).} To enable this, we used a snow machine \textred{(Table~\ref{tab:equipment})} and moved it around manually to cover the entire flight area with falling snow.  This experiment had a success rate of $75\%$, with 15~successes and 5~failures over 20~trials.

\subsubsection*{\textcolor{black}{Dense fog}}
Low-visibility conditions such as fog, smoke, and/or dust are common in natural environments and particularly relevant in disaster search and rescue operations such as earthquakes, forest fires, cave extraction and mountain rescue operations where reliable navigation under such is the difference between life and death. Big, bulky and expensive LIDARs are semi-operational \cite{Alexis_Darpa2022} in these conditions but limit the size of the robot to traverse through tight spaces often found in collapsed 
structures. With our fog experiments, we showcased that ultrasound sensors coupled to our approach provide a promising solution for navigation in smoke and fog (Fig.~\ref{fig:indoor_nav_image}\textcolor{red}{D}). 
The obstacles used were a combination of boxes, with dimensions in the range of $\{46,67\}\times\{115,138\}\times\{19,25\}$~cm (length $\times$ height $\times$ thickness).
To create a smoky/foggy environment we utilized two fog machines \textred{(Table~\ref{tab:equipment})} to cover the entire obstacle course.
 We used enough fog such that the visibility of the camera-based sensors would be less than 0.75~m. The success rate of this experiment was $90\%$, with 18~successes and \textred{two} failures.

\subsubsection*{Low-light}
Low-light conditions pose several challenges to vision-based sensors, including increased noise, long exposure times, motion blur, and underexposed images. These factors degrade the accuracy and \textred{often} cause a complete failure of vision-based perception algorithms. \textred{Although} active illumination can mitigate some of these effects, it often incurs substantial power consumption. Furthermore, in dusty/hazy conditions, active illumination essentially blinds the sensors with glare that requires expensive custom optics \cite{skydioX10}. We demonstrated the value of using ultrasound sensors in low-light (0.2~lux) conditions. We utilized three blue-colored translucent play tunnels ($\diameter$: 46~cm, height: 122~cm, \textred{Table~\ref{tab:equipment}})
arranged randomly in the environment such that the robot cannot travel in a straight line without dodging at least two obstacles (Fig.~\ref{fig:indoor_nav_image}\textcolor{red}{E}). The success rate of this experiment was $100\%$, with 20~successes.

\subsubsection*{Composite}
Different obstacles exhibit distinct acoustic reflection characteristics. For instance, cardboard boxes produce stronger reflections (high PSNR) but are only detectable within a limited angular range due to sharp corner falloff \cite{martinez_using_2003}. In contrast, cylindrical obstacles yield weaker reflections (low PSNR) but can be perceived from a broader range of directions. To emulate realistic environments with diverse obstacle geometries, we constructed a cluttered scene composed of six obstacles,
with dimensions in the range of $\{25,46\}\times\{121,138\}\times\{19,45\}$~cm (length $\times$ height $\times$ thickness).
These obstacles were distributed in the scene so that any arbitrary straight-line motion through this set-up would result in a collision (Fig.~\ref{fig:highspeed_batdeck_comparison}\textcolor{red}{A(i)}). The success rate for this experiment was $69.57\%$, with 16~successes and \textred{five} fails.

\subsubsection*{\textred{Comparison against other ultrasound navigation solutions}}
We evaluated our approach against BatDeck\cite{muller_batdeck_2024}, which employed the identical ICU30201 sensor. For comparative testing, we coupled the ICU30201 with a $45^\circ$ acoustic horn, approximating BatDeck's $55^\circ$ \textred{Field of View (FoV)} horn. Our setup featured a single front-facing sensor mounted on a physical noise reduction shield. We implemented BatDeck's low-pass filtering algorithm alongside its obstacle detection and avoidance logic. This system modulates forward velocity linearly to zero based on proximity to the nearest detected obstacle, while simultaneously applying angular velocity around the yaw axis until the obstacle moves out of sensor view. Testing in our composite indoor environment revealed \textred{substantial} performance differences. BatDeck's approach (Fig.~\ref{fig:highspeed_batdeck_comparison}\textcolor{red}{A(ii)}), which randomly alternates turning direction every ten seconds, successfully navigated the obstacle course only once in 17~trials, with \textred{five} direct obstacle crashes and 11~net collisions. In contrast, our method (Fig.~\ref{fig:highspeed_batdeck_comparison}\textcolor{red}{A(i)}) successfully completed the course 13~times, with only \textred{three} net collisions and \textred{seven} crashes in identical testing conditions, demonstrating substantially improved obstacle avoidance capabilities.

\subsubsection*{Speed runs}
In this experiment, we evaluated obstacle avoidance performance during flights at different speeds (see Fig.~\ref{fig:highspeed_batdeck_comparison}\textcolor{red}{B}). The environment consisted of three obstacles, 
with dimensions in the range of $\{25,69\}\times\{121,170\}\times\{19,25\}$~cm (length $\times$ height $\times$ thickness)\textred{.} 
The scene was intentionally kept sparse to allow the aerial robot to accelerate to the desired speeds. We experimented with three different target forward speeds: $1~\mathrm{m\,s^{-1}}$, $1.5~\mathrm{m\,s^{-1}}$, and $2~\mathrm{m\,s^{-1}}$ and achieved success rates of 100\%, 81.82\%, and 72.73\% respectively. As expected, the success rate dropped with increasing speed, \textred{as} other conditions, including navigation parameters and scene layout, were held constant.

\subsubsection*{\textcolor{black}{3D obstacle avoidance}}
\begin{figure}
    \centering
    \includegraphics[width=0.95\linewidth]{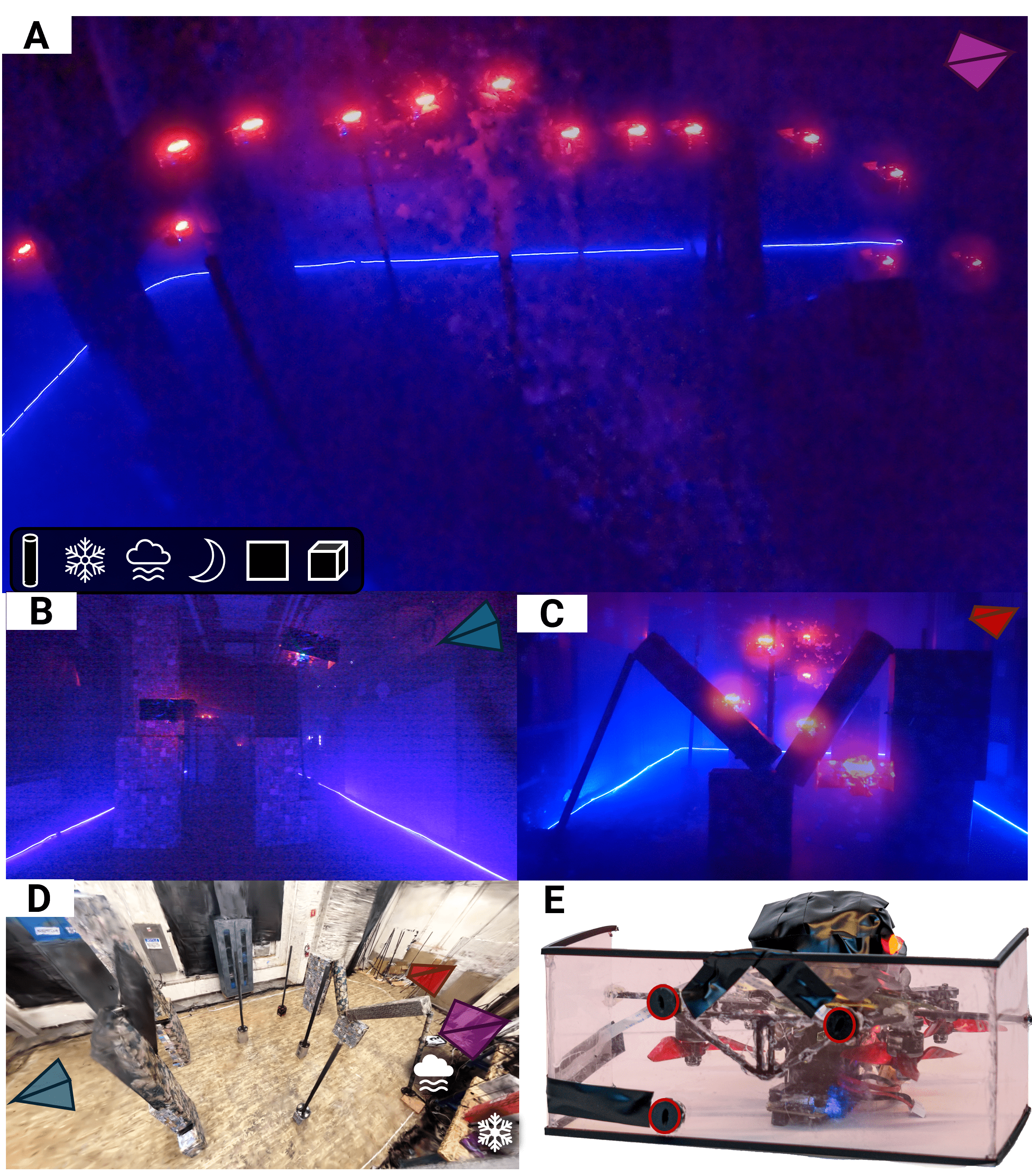}
    \caption{\textbf{Aerial robot navigation through a complex 3D indoor scene with smoke, snow, textureless and thin obstacles.} (\textbf{A}) Top view (robot motion is right to left), (\textbf{B}) Back view (robot motion is towards camera), (\textbf{C}) Front view (robot motion is away from camera), (\textbf{D}) 3D digital twin created using VizFlyt \cite{vizflyt2025} during well-lit conditions; The camera frustums show the respective camera locations (\textcolor[RGB]{160, 43, 147}{top}, \textcolor[RGB]{21, 96, 130}{back} and \textcolor[RGB]{192, 0, 0}{front}). The fog and snow icons in (\textbf{D}) show the locations of fog and snow machines, respectively. \textred{(\textbf{E}) Tri-ultrasound setup is used in the 3D experiment to enable trilateration for 3D obstacle avoidance. \textcolor[RGB]{192, 0, 0}{Red} ellipses show the ultrasound sensors.} Opacity shows the time progression.}
    \label{fig:indoor_3dnav_image}
\end{figure}

\textcolor{black}{
Since aerial robots operate in three-dimensional environments, both horizontal and vertical obstacle dodging capabilities are essential for practical navigation. In the previous set of experiments, we demonstrated only lateral movements for obstacle avoidance to demonstrate the efficacy of our approach. To assess practical usability in the real world, we constructed a complex 3D scene where motion in both horizontal and vertical directions is essential for dodging in this experiment. The indoor scene consists of horizontal, vertical, and oblique obstacles made using PVC poles (dimensions: \{4, 6\}$\diameter \times$192~cm), cardboard boxes with dimensions in the range of $\{24,67\}\times\{62,129\}\times\{19,25\}$~cm (length $\times$ height $\times$ thickness), a cardboard cylinder (25$\diameter \times 121.5$~cm) and a cardboard with a matte black featureless sticker stuck on it ($59\times100 \times0.5$~cm) arranged to create a complex 3D obstacle course \textred{(}Fig.~\ref{fig:indoor_3dnav_image}\textred{A-D}).  
Initial robot positions were randomized while ensuring each trajectory necessitated 3D maneuvers. The perception and planning approach for 3D navigation is detailed in the Materials and Methods (Extension to 3D) section. We evaluated performance under three combined weather conditions: low light, dense fog, and snow (Supplementary \textred{Movie} \textred{S2}). The success rate for this experiment was 72.7\% with 16 successes and \textred{six} failures.}

\subsection*{Outdoor experiments}

\begin{figure}
    \centering
    \includegraphics[width=0.75\linewidth]{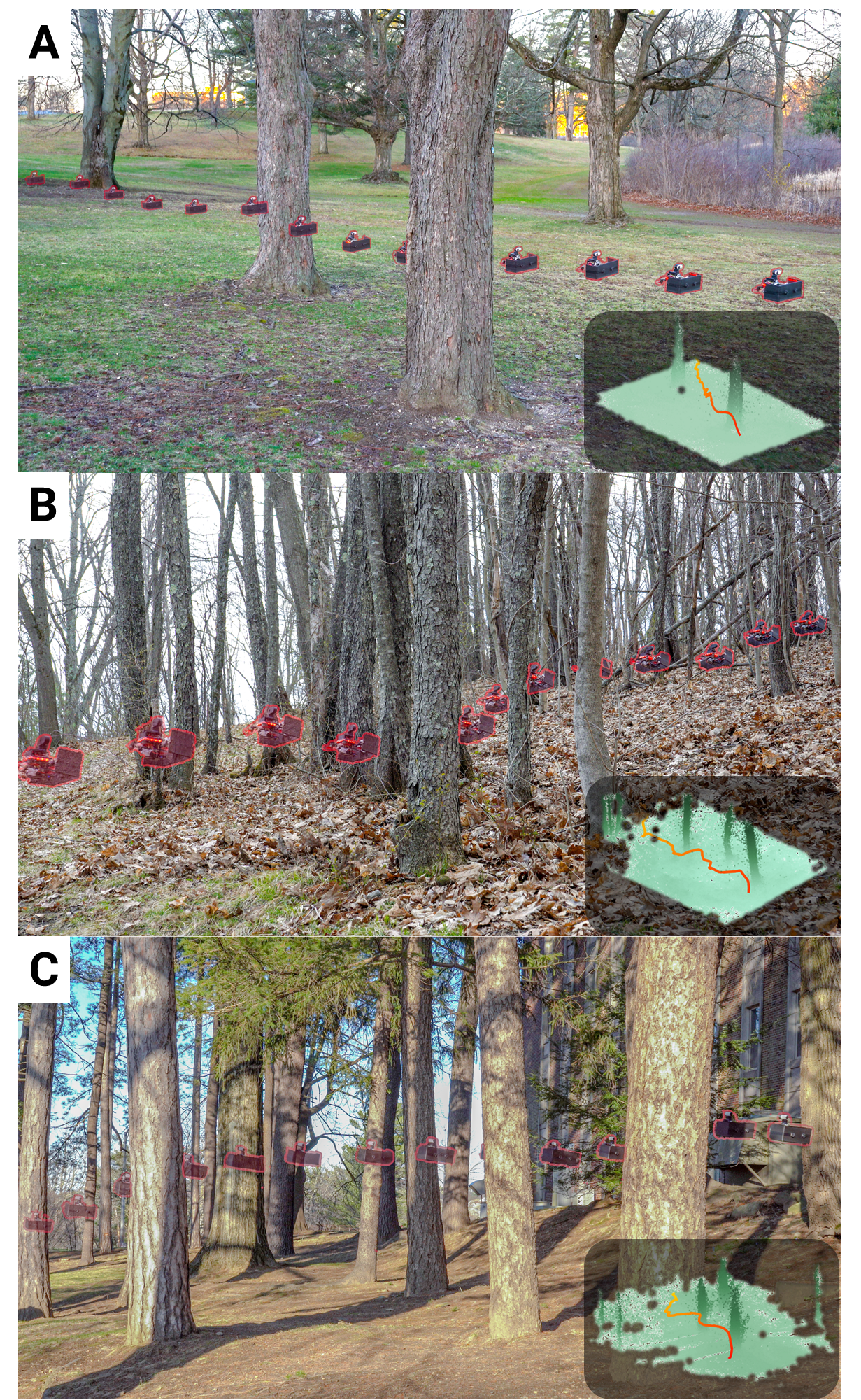}
    \caption{(\textbf{A--C}) \textbf{Aerial robot navigation in outdoor forest environments.} \textred{Insets} show 3D reconstructions from onboard camera footage using Structure-from-Motion (SfM) to serve as a reference. Red to yellow and opacity increase show time progression. The robot is highlighted in red for the sake of clarity.}
    \label{fig:outdoor_images}
\end{figure}

\textred{Although} indoor experiments offer more controlled conditions, outdoor scenarios introduce a unique set of challenges. \textcolor{black}{Trees typically reflect a smaller portion of the transmitted ultrasonic signal compared to smooth indoor obstacles, leading to a lower signal-to-noise ratio}. Moreover, the shapes and structures of trees can vary \textred{substantially} between locations, and additional environmental factors such as wind further complicate perception and navigation. We extensively evaluated our approach across three different forest environments with low, medium, and high tree densities. The aerial robot consistently avoided obstacles, achieving a success rate of $90.9\%$, $77.3\%$, and $85.7\%$ across all three scenarios. We mounted an Insta360\texttrademark{} camera onboard to record first-person RGB images, which were used with the COLMAP Structure-from-Motion (SfM) pipeline \cite{schoenberger2016sfm} to reconstruct and visualize the robot trajectories during obstacle avoidance (Fig.~\ref{fig:outdoor_images}).

\subsection*{Noise reduction}
\subsubsection*{Propeller induced noise}
Vertical lift aerial robots generate substantial noise that can disrupt acoustic sensors, including ultrasonic ones.  This potentially reduces the PSNR (peak signal-to-noise ratio), rendering obstacles undetectable at the noise levels we encounter. We analyzed propeller-induced noise and its \textred{effects} on signal detection through two methodical tests.

We configured the ICU30201 ultrasonic sensor in receive-only mode and recorded the noise while operating our propulsion system at varying throttle settings. To comprehensively assess noise variation across different propeller dimensions, we tested multiple propeller sizes while maintaining a fixed sensor position relative to the propeller tip. We analyze the \textred{influence} of a single propeller motor combination. The sensor was positioned facing away from the propeller's tip at a distance of 32~mm in the propeller plane. We computed the RMS of the received signal amplitudes, with results presented in Fig.~\ref{fig:physical_noise_reduction_ablation}\textcolor{red}{C}. \textcolor{black}{For example, for a $60~mm~\diameter$ PVC pole placed at 1~m distance directly in front of the ICU30201 sensor, with propellers generating hover thrust ($0.46~$kg), the Peak Signal-to-Noise Ratio (PSNR) is $-4.9$~dB.}

\subsubsection*{Physical noise reduction}

From Fig.~\ref{fig:physical_noise_reduction_ablation}\textcolor{red}{C}, propellers generate substantial noise that interferes with ultrasonic sensor performance. We demonstrate that physical shields provide an effective mitigation strategy. This section analyzes shield efficacy across different dimensions and materials.
Sound waves can bypass obstacles through diffraction or transmission, with larger shields generally providing better blocking. However, robot design constraints necessitate optimizing shield size to minimize weight and volume requirements. We first investigated noise reduction performance using shields of varying dimensions\textcolor{black}{, tested with 3$\times$ 76.2~mm propellers (HQProp DP 3$\times$4$\times$3")}. For simplicity, we maintained a planar shield geometry using 4.5~mm polystyrene foam for ease of fabrication. The results of our dimensional ablation study are presented in Figs.~\ref{fig:physical_noise_reduction_ablation}\textcolor{red}{E}, \textcolor{red}{F}, \textcolor{red}{G}.

We subsequently evaluated material effects on noise reduction performance, as shown in Fig.~\ref{fig:physical_noise_reduction_ablation}\textcolor{red}{D}. Our material comparison included three options: 4.5~mm polystyrene foam, 0.5~mm plastic, and a composite shield combining 0.5~mm plastic with 4~mm Aggsound foam (Table.~\ref{tab:equipment}). Throughout these material tests, we maintained consistent shield dimensions, propeller-to-sensor distances and \textcolor{black}{tested with 3$\times$ 76.2~mm propellers (HQProp DP 3$\times$4$\times$3')}. The complete experimental setup is illustrated in Fig.~\ref{fig:physical_noise_reduction_ablation}\textcolor{red}{B}. 
\textcolor{black}{As seen in Fig.~\ref{fig:physical_noise_reduction_ablation}, the lowest shield height with the most noise reduction was 63.5~mm, and all decreases in noise with \textcolor{black}{larger shields} were minimal (we call this the ``optimal height'').}

\subsubsection*{Comparison against denoising methods}

The efficacy of the Saranga network was evaluated against four classical techniques for noise reduction and edge detection. We utilized Total Variation Denoiser (L1 regularizer, $\lambda=1$) \cite{chambolle2004algorithm}, Gaussian Blur ($5\times 5$ kernel)\cite{davies_machine_1990}, Total Variation Denoiser (L1 regularizer, $\lambda=1$) combined with a Savitzky-Golay (SG) filter (window size 11, 7th order) \cite{savitzky1964smoothing}, and Two-Dimensional Least Mean Squares (TDLMS) filtering ($9\times 9$ kernel)\cite{hadhoud1988two}, followed by a Sobel filter to extract edges corresponding to obstacle locations. Using the algorithm described in Algorithm \ref{alg:obstacle_detection_3d} \textred{in the Supplementary Methods}, we estimated the location of obstacles.

Synthetic motion data was generated for a single point obstacle under fixed signal strength and convolved to produce an ideal echo response. To simulate realistic conditions, propeller-induced noise at varying intensities was added to the generated data (see Fig.~\ref{fig:datagen_input}). Ground truth was used to compute the accurate $(\mathcal{O}_x, \mathcal{O}_y)$ Cartesian position of the obstacle. Since the edge maps produced by Sobel edge detection and Saranga could be offset relative to the ground truth, we applied a compensation offset prior to metric computation. We evaluated the Root Mean Squared Error (RMSE) of obstacle position, Structural Similarity Index (SSIM), and Mean Squared Error (MSE/L2) across various noise levels (characterized by PSNR) (Figs.~\ref{fig:classical_compare}\textcolor{red}{G(i)-G(iii)}). \textred{Additionally, we measured the runtime and energy consumption for a single inference or execution of each method, as summarized in the table in Fig.~\ref{fig:classical_compare}.}

To validate \textcolor{black}{real world} performance, we mounted our system on an aerial robot and moved a thin pole in the field of view. Thin poles are weak reflectors and easily masked by noise. We qualitatively assessed the denoising and edge detection performance of each method on this real data, again under varying PSNR conditions (Figs.~\ref{fig:classical_compare}\textcolor{red}{B-F}).

We leveraged the Google Coral Tensor Processing Unit (TPU) to accelerate inference of the Saranga network. We implemented the classical filters such as TDLMS, Gaussian, Savitzky-Golay and Sobel using SciPy. For Total Variation denoising filter, we used the Scikit-image library.

\subsection*{\textcolor{black}{Comparison against traditional sensors}}
\textcolor{black}{We evaluated the efficacy of ultrasound as a sensing modality for aerial robots against four common sensors. These sensors include a uRAD automotive RADAR, a CygBot D1 LIDAR (3D mode), and an Intel RealSense D435 camera (with and without active illumination). An obstacle was positioned one meter away from the sensor and the accuracy was computed. Accuracy is defined as $1-\mathbb{E}(\vert\hat{D}_i-\tilde{D}_i\vert)$ where, $\hat{D}, \tilde{D}$ represents the ground truth and measured depth respectively, $i$ indexes over all sensor measurements, $\mathbb{E}$ is the expectation operator. For undefined/zero measurements, we use a maximum error of 1. These results are summarized in Figs.~\ref{fig:sensor_comparison_complete}\textcolor{red}{B, C}.}

\textcolor{black}{We also ablated on various ecological conditions (snow, fog, low-light) and obstacle material types (boxes, black featureless sheet, thin transparent film and glass). Low-light \textred{conditions have} an ambient light intensity of 0.1 lux, in a room with no lights and covered windows. All other scenes were well-lit, measured around 134 lux.  We used varied obstacles to test each sensor. For varying ecological conditions (snow, fog, low-light), we used stacked cardboard boxes with a mossy stone wallpaper as the obstacle (same as indoor flight experiments). The snow and fog are artificially produced using the same machines used in our indoor experiments. Further, to test how the sensors performed on featureless objects, we used a foamcore sheet with a matte black wallpaper (Table.~\ref{tab:equipment}) stuck on it. The thin transparent film is of 0.75mil (0.02mm) thickness, the same used in our transparent obstacles indoor flight experiment. Finally, a 3mm thick glass pane was used to compare the efficacy on a rigid transparent object. We refer the reader to the supplementary material (Fig. \ref{fig:obstacle_images}) for images of these materials. Sonar is able to work with an average of 89.3\% accuracy for all environments/obstacles (snow, fog, darkness, featureless black, glass, and thin plastic film obstacles). \textred{Although} RADAR is often used in harsh conditions, and performs well in many scenarios, it was unable to detect the thin plastic sheet, demonstrating its dependence on material and thickness. A detailed discussion of the results is presented in the next section.}

\subsubsection*{\textcolor{black}{Doppler effect}}
\textcolor{black}{Given that the ultrasound sensor is on a moving robot, it is useful to analyze the \textred{influence} of the Doppler effect on the received signal strength. If an aerial robot moves head-on toward a stationary obstacle at velocity $v$ and the speed of sound is given by $c_s$, the observed frequency $f_o$ would be \cite{neipp2003analysis},
\begin{equation}
    f_o = \frac{c_s + v}{c_s-v}f
\end{equation}
For a speed of $6.3~\mathrm{m\,s^{-1}}$, the frequency shift $f_o - f$ is approximately $1.983~\mathrm{kHz}$. This represents about half the sensor bandwidth (measured at 3.98~kHz) for our ultrasound sensor operating at 53~kHz \cite{tdk_icu30201}. At this speed, the received power would be reduced by half due to the Doppler effect alone. However, this is not a concern since our experiments operate at a maximum speed of $2~\mathrm{m\,s^{-1}}$. In future work, we hope to utilize the Doppler shift as an informational cue for improving obstacle localization accuracy at higher speeds.}

\section*{DISCUSSION}
\textcolor{black}{Current perception systems typically rely on information-rich sensors like cameras or LIDAR, with high resolution and accuracy. For search and rescue tasks, cost and sensor applicability under varied ecological conditions are pivotal. Although deep learning methods can improve operational domain based on standard sensors\cite{Chen2018LearningTS}, the problem is still ill-posed for common sensors like cameras and LIDARs in visually degraded conditions like snow, fog/smoke, and low-light. To this end, we evaluate commonly available sensors across obstacle types to aid obstacle avoidance in the wild. }

We compare the physical characteristics such as size, weight, power, sensing range, and cost for sonar (ultrasound), RADAR, LIDAR, and Intel RealSense D435 (Fig.~\ref{fig:sensor_comparison_complete}\textcolor{red}{B}). Sonar (ICU30201) scored best in power, size, weight, and cost, \textred{whereas} RADAR (uRAD Automotive HPA) had the highest range. We then empirically measured the accuracy of these sensors under adverse conditions, including snow, fog, darkness, featureless black objects, plastic film, and glass (Fig.~\ref{fig:sensor_comparison_complete}\textcolor{red}{C}). As expected, sonar detected objects in all scenarios, including complete darkness (0 lux) and black-featureless obstacles, with RADAR coming in a close second, working in most environments except the plastic film, since plastics reflect radio waves poorly\cite{radome_kumar2021}. The LIDAR showed degraded performance in fog due to light scattering, and failed completely on featureless black matte objects, glass and transparent film, either due to low reflections or completely passing through the object. RealSense D435 (laser ON/OFF) failed under fog and glass, with degraded accuracy in snow. Contrary to expectations, RealSense D435 detected featureless black objects, likely due to well-tuned feature matching and hole-filling algorithms. The RealSense D435 detected the thin plastic film with degraded accuracy, likely because wrinkles scattered the laser beam. With the laser OFF, RealSense was able to detect the transparent film with higher accuracy compared to laser ON. We believe this is because with the emitter ON, the RealSense feature matches objects behind the thin film rather than on the film. \\
It is important to point out that RADARs can be a good sensor choice for long-range high-speed autonomy if power, size and cost budget allow for it. In extreme situations of low power, low cost and smaller size, ultrasound would be better. The much lower power footprint of sonar (2833$\times$) is crucial for search and rescue missions where critical flight times would be the difference between life and death for survivors. \textred{When designing a parsimonious robot, constrained by size, weight, area, and power (SWAP), and pushing for end-to-end autonomy in diverse environments, it is crucial to select a sensor perfectly crafted for the problem at hand.} 

\begin{figure}
    \centering
    \includegraphics[width=\linewidth]{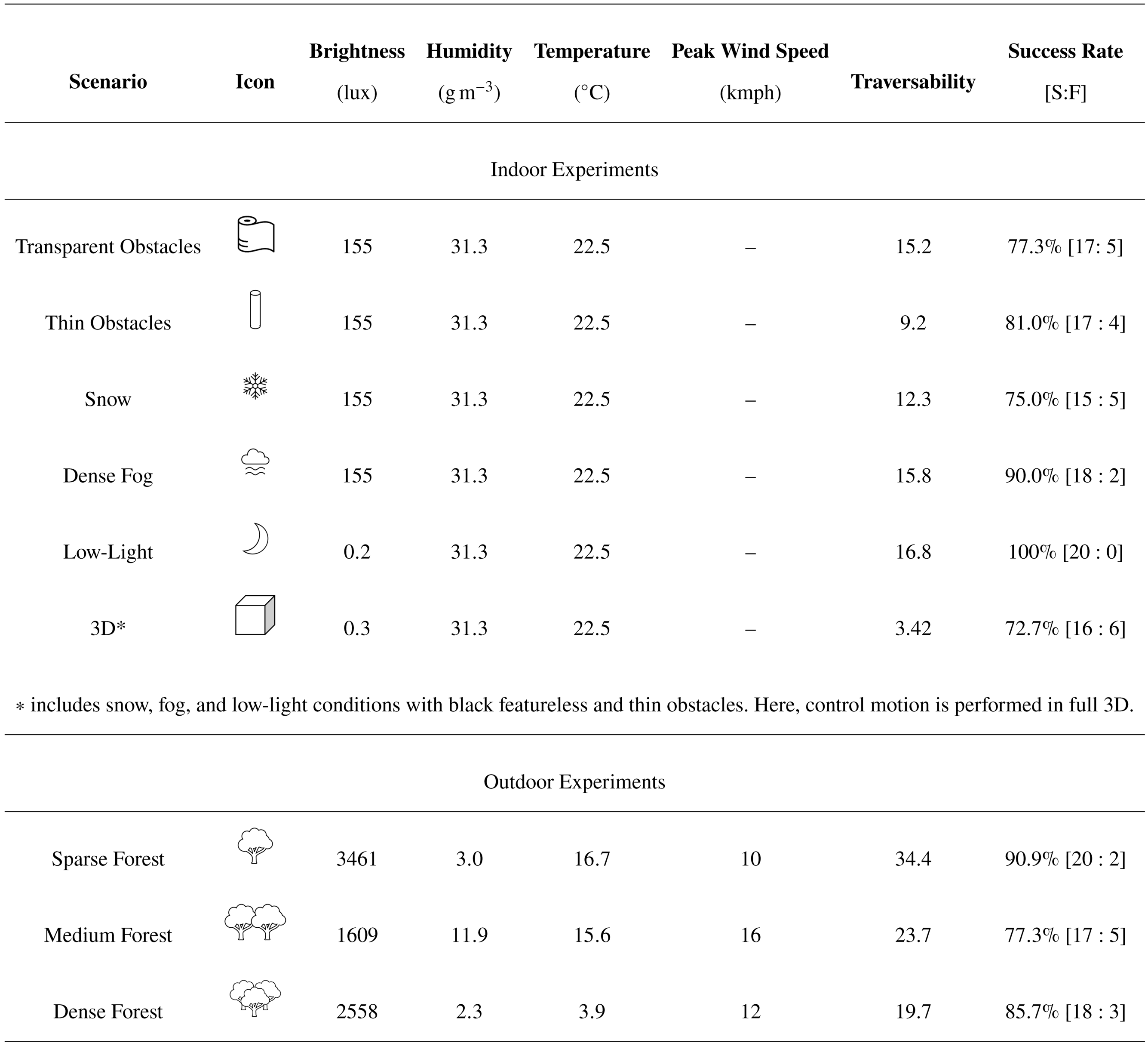}
    \caption{\textbf{Tabulation of navigation performance.} Variation of performance across diverse challenging scenarios with varying obstacles and environmental conditions.}
    \label{fig:success_rates}
\end{figure}

In this work, we introduce an ultrasound-based Saranga framework for a resource-constrained, lightweight aerial robot, capable of navigating with minimal computational power in any environment. Our approach performs robustly in challenging conditions like snow, \textcolor{black}{fog}, low-light, and against objects with high transparency (Figs.~\ref{fig:success_rates} and \ref{fig:indoor_nav_image}). With just two ultrasound sensors, we are able to overcome many of the limitations of traditional vision-based and other popular sensing technologies.

As shown in Fig.~\ref{fig:classical_compare}, \textred{although} classical techniques can denoise and extract edges, they perform reliably in carefully curated scenarios, with well-tuned filters. They tend to produce spurious detections (Figs.~\ref{fig:classical_compare}\textcolor{red}{B--E}) or miss obstacles entirely when edge detection thresholds are increased. Moreover, methods like \textred{Total Variation denoising (TV-L1)} are computationally intensive due to the lack of hardware acceleration, resulting in higher energy and runtime costs (Fig.~\ref{fig:classical_compare}). In contrast, Saranga benefits from hardware acceleration on a Google Coral Tensor Processing Unit (TPU), retains obstacle range information more accurately (Fig.~\ref{fig:classical_compare}\textcolor{red}{F}), and delivers notable improvements in impulse detection and obstacle localization accuracy (Figs.~\ref{fig:classical_compare}\textcolor{red}{G(i)--G(iii)}).

As seen in Fig.~\ref{fig:physical_noise_reduction_ablation}\textcolor{red}{C}, \textcolor{black}{propeller noise increases with generated thrust. Heavier aerial robots would therefore experience higher noise levels for identical propeller sizes, though larger propellers produce substantially lower noise at the same thrust requirement.}

Additionally, signal power drastically drops with increasing distance. For example, signal strength drops by $17$~dB when an obstacle is moved from $0.5$~m to $2.0$~m. \textred{To increase the detection range, either the transmit pressure should increase, or the noise levels must drop.} The maximum transmit pressure has two possible limiting factors: the linear case, multiplying the transmit sensitivity of the transducer with the maximum available transmit voltage, or the non-linear case, where third-order mechanical non-linearity in the piezoelectric membrane stiffens the membrane during large amplitude deflection. In the case of ICU-30201, the transmit voltage is 20~V$_{\mathrm{pp}}$ per electrode, and \textcolor{black}{Piezoelectric Micromachined Ultrasonic Transducers (pMUTs)} of similar design operate near or at the non-linear limit \cite{przybyla2023mass, przybyla2015}. The transmit pressure cannot be increased further by simply increasing the transmit voltage due to the mechanical non-linearity present in the \textred{Micro-Electro-Mechanical Systems} (MEMS) transducer element.\\
In case the sensor characteristics cannot change, having a lower FoV can help with detecting objects that are far away. But this can come at the cost of colliding with obstacles that are close to the side of an aerial robot. The shapes and sizes of obstacles have an effect on the signal strength of the received echoes\textcolor{black}{, resulting in the varying success rates across the experiments as environments and obstacles change over trials}. A PVC pipe reflects at half the intensity compared to a large flat surface.
But although the planar objects, like a box, reflect ultrasonic signals well, they tend to deflect the transmitted beam away from the robot, especially while facing an edge/corner. This resulted in a reduced success rate for experiments with boxes (Figs. \ref{fig:success_rates}, \ref{fig:indoor_nav_image}\textcolor{red}{C, D}). {Snow} and \textcolor{black}{fog} do not seem to affect our perception system in any measurable way.
Thin cylindrical obstacles reflect at a lower intensity but typically can be avoided from different approach angles. 
Thin obstacles are detected at a very close range (<40~cm), which does not give enough time for the navigation system to react for dodging. This resulted in lower success rates with thin poles (Fig.~\ref{fig:success_rates}). Additionally, because of the wide field of view (Fig.~\ref{fig:beam_pattern_distance}) and closely packed obstacles (<0.5~m gap between some of them, Fig.~\ref{fig:indoor_nav_image}\textcolor{red}{B}), the robot continued to dodge poles sideways, and in this process the rear of the robot could hit another pole it had just passed.\\
With trees, detections occurred closer than 40~cm due to uneven surfaces in a few trials, which did not give enough time for the navigation stack to react. Additionally, wind-induced drifts resulted in collisions; even after dodging, wind could sometimes push the robot back into the obstacle. Our aerial robot struggles to avoid thin branches in natural environments due to weak reflected signals. Although this remains a limitation, catastrophic collisions were prevented by the physical shield, which not only attenuated propeller-induced noise but also provided mechanical protection. Despite the shield's effectiveness, further reducing noise levels remains valuable, as it could extend sensing range and improve obstacle detection in low-reflectance, high-noise scenarios.

\textcolor{black}{\textred{Although} the current work focused on denoising and making ultrasonic sensors viable on aerial robots, there is scope for improvement in the perception and planning stack. There is a risk of collision with other obstacles while reactively dodging an obstacle. Maintaining a local map of obstacles may be helpful in avoiding such failure scenarios, which we hope to explore in the future. Additionally, the current perception and navigation method relies on tuning several parameters, including the detection threshold ($\tau_c$), desired velocity ($V_d$), hysteresis ($\delta$), repulsion gains ($K$), and maximum lateral distance ($\delta_{max}$). The robot's avoidance behavior and clearance margins are highly sensitive to these tuning parameters. For example, increasing the detection threshold also increased the likelihood of missing detections, \textred{whereas} decreasing the detection threshold improved detection performance for thin obstacles and outdoor uneven obstacles (trees). \textred{Although} increasing the repulsion gains increased clearance to obstacles, it also resulted in suboptimal trajectories involving excessive lateral movement (Fig.~\ref{fig:indoor_nav_image}\textcolor{red}{E}). Higher hysteresis thresholds helped dodge fluttering obstacles like transparent film (Fig.~\ref{fig:indoor_nav_image}\textcolor{red}{A}) by keeping control actions relatively smooth and decisive in one direction, avoiding rapid left-right velocity commands caused by continuously fluctuating obstacle localization due to film vibrations induced by propeller downwash. At higher speeds, increased sensitivity to obstacle detections (i.e., lower confidence thresholds) and higher control gains were critical for timely dodging but can result in oscillatory motions. These tradeoffs could potentially be mitigated by an adaptive or learned planning strategy, which will be explored in future work.}

The current \textcolor{black}{Mean Square Error (MSE)} loss function penalizes denoising errors uniformly across the entire time horizon. However, our application demands higher accuracy in the most recent time frame. Addressing this temporal weighting will be a focus of future work.

\textcolor{black}{The robot used in this work uses over 100~W of power at hover, indicating that the savings obtained from our sensing stack (1.2~mW) are not \textred{substantially} beneficial for the current prototype. It is important to note that a future iteration with smaller quadrotors like the Crazyflie with the GAP9 AI shield \cite{muller2024gap9shield} can \textred{noticably} benefit from the power savings of our sensing stack. Furthermore, TPU-compatible microcontrollers are becoming increasingly common and lightweight. For example, the Google Coral$^\text{TM}$ Micro\cite{coral_micro} weighs only 10~grams while integrating the same Edge TPU as the Google Coral$^\text{TM}$ Mini\cite{coral_mini} used in this work, reducing compute weight from 26~grams to 10~grams. Also, microcontrollers with on-chip neural network accelerators (e.g., STM32N6\cite{stm32n6}) are emerging, further reducing both computational power requirements and physical footprint. Future work will explore integrating these lighter compute solutions to enable deployment on miniature quadrotors ($<50$~mm diagonal length). In a real search and rescue mission, a few more seconds of precious flight time could be the difference between life and death for a survivor.}

\textred{Although} a UNet architecture performs effectively on our compute platform (Google Coral Mini), the computational load could be further reduced by exploring lightweight recursive architectures such as \textsc{GRU}s or \textsc{LSTM}s. If the inference cost is sufficiently minimized, the denoiser could eventually be deployed directly on the sensor package itself, enabling more compact and integrated designs.

A detailed study on the effect of various noise sources across broadband and narrowband ultrasonic sensors remains lacking. \textred{Although} our analysis focused on four propellers relevant to the $<500~\mathrm{g}$ aerial robot category (Movie.~\ref{fig:banner}), extending this investigation to a broader range of propellers and frequency bands could provide deeper insights. Such a study would not only enhance understanding of sensor-propeller interactions but also serve as a valuable design reference for the wider scientific community.

When compared against BatDeck \cite{muller_batdeck_2024}, we demonstrate a substantial improvement in navigating through an obstacle course while experiencing fewer collisions. We make informed navigation decisions with full knowledge of the obstacle's position, whereas BatDeck \textcolor{black}{employs a reactive strategy that scales velocity and yaw rate based solely on distance to the closest detected obstacle. BatDeck randomly alternates turn direction every 10 seconds and cannot distinguish obstacle azimuth or elevation, limiting its ability to make targeted avoidance maneuvers} (Fig.~\ref{fig:highspeed_batdeck_comparison}\textcolor{red}{A(ii)}). Additionally, we observed that having a higher FoV sensor ($140\times 57^\circ$) proved to be helpful in two scenarios - during \textred{Extended Kalman Filter (EKF)} failures or due to wind, when the robot drifted sideways, the robot would continue to dodge the obstacles that appeared from the sides. Additionally, we maintained good clearance sideways for all our experiments (Fig.~\ref{fig:indoor_nav_image}).

\section*{MATERIALS AND METHODS}
\label{sec:materials_methods}
\subsection*{Problem overview}
\label{sec:overview}
We aimed to navigate an aerial robot towards a goal direction while avoiding obstacles of unknown shape, size, location, and material under unknown harsh and \textcolor{black}{varied} environmental conditions. Our robot was equipped with two low-power MEMS ultrasonic chirp sensors (called left and right). \textcolor{black}{During} each measurement cycle, the left sensor transmitted a short ultrasound pulse at the resonant frequency of the MEMS element, after which both sensors received the echoes (a chosen amount of time to ``listen'' to the echoes). Given echo measurements from both the sensors, we could bilaterate the obstacle to find its relative position to the sensor suite using simple trigonometry \cite{nagashima_1992}. However, there were few major problems that needed to be rectified to be an effective method for aerial robot navigation: propellers generate a high amount of \textcolor{black}{wide-band} noise including ultrasound frequencies, the transmitted sound pulse has a relatively low amplitude compared to possible interference sources, and echo reflection strength can vary \textred{substantially} at every measurement step depending on the relative orientation of the obstacle, obstacle material, and interferences. \textcolor{black}{To tackle these problems and enable robust obstacle detection, we proposed a two-pronged approach:} deep-learning-based noise reduction and processing using Saranga, and physical noise reduction. We will explain each of these methods in the subsequent sections. Before we delve into the details, we will first introduce the notations that will be used in the remainder of the paper.

The echo signal received by each sensor is an amplitude waveform \textcolor{black}{captured during each transmit/receive cycle.} The \textcolor{black}{time of flight} of an echo from an object at range $R$ from the sensor is given by $\frac{2R}{c_s}$, where $c_s$ is the speed of sound in air, which is $343~\mathrm{m\,s^{-1}}$ at 20$^\circ$~C. We \textred{indexed} measurements in each measurement cycle with $r$ referring to range. Ideally, each obstacle resulted in a strong peak at a time corresponding to the distance $R$ (given by $\frac{2R}{c_s}$). Detecting peaks in each received echo from two sensors and correlating them to find correspondences to compute the time difference enabled obstacle localization (since both sensors \textred{received} an echo from the same obstacle at slightly different times due to different path lengths).

However, since the echoes can reach the sensor from multiple paths, they were plagued by attenuation and noise from the interferences and propeller-induced noise. This \textred{made} the peaks unstable and hard to correlate between the two sensors. To counteract these issues, we used the time history of the signals $t$ from both sensors, i.e., the stacked echo signal. To find accurate peak correspondences, we needed to denoise the signal, which was difficult due to the multitude of effects that are difficult to model. Serendipitously, the noise is uncorrelated in time $t$. Imagine an echo image $\mathcal{E} \in \mathbb{R}_+^{M\times N}$, where each row corresponds to the index in time $t$ and the columns represent amplitude waveform \textcolor{black}{corresponding to obstacle(s) in range $\{r\}$}. This echo image is analogous to the M-mode image in ultrasound imaging\cite{fraser_concise_2022, edler_use_2004}. Essentially, we wanted to ``look'' for temporally consistent data and ignore the temporally inconsistent noise. To this end, we \textred{proposed} a neural network-based denoising and signal-processing solution called Saranga, which is explained next.

\subsection*{Saranga}
\label{sec:saranga}

We propose Saranga, a deep neural network that takes as input the raw echo image $\mathcal{E}$ and outputs the denoised echo image $\hat{\mathcal{E}}$, which looks like the ideal impulse response from a point obstacle, i.e., the history plot looks like a line (edge in computer vision, see  Fig.~\ref{fig:sensor_comparison_complete}\textcolor{red}{D}). Since it is arduous, time- and cost-prohibitive, and borderline impossible to obtain a ground truth map for ultrasound echo image $\mathcal{E}$ and its ideal version, we first mathematically modeled the ideal echo image in simulation and then injected noise from real samples for a seamless simulation to real (sim2real) transfer. At the heart of this process is the data generation pipeline as described next. \textcolor{black}{In the context of our work, sim2real transfer \textred{referred} to generalization to real scenes with a robot in motion, where obstacles are detected without any real data of obstacle motion used for training. However, we used a very small amount of real noise data to facilitate this generalization. This is similar in spirit to calibration regimes in literature \cite{Kaufmann2023,wu2019}.}

\subsubsection*{Data generation}
\begin{figure}
    \centering
    \includegraphics[width=\linewidth]{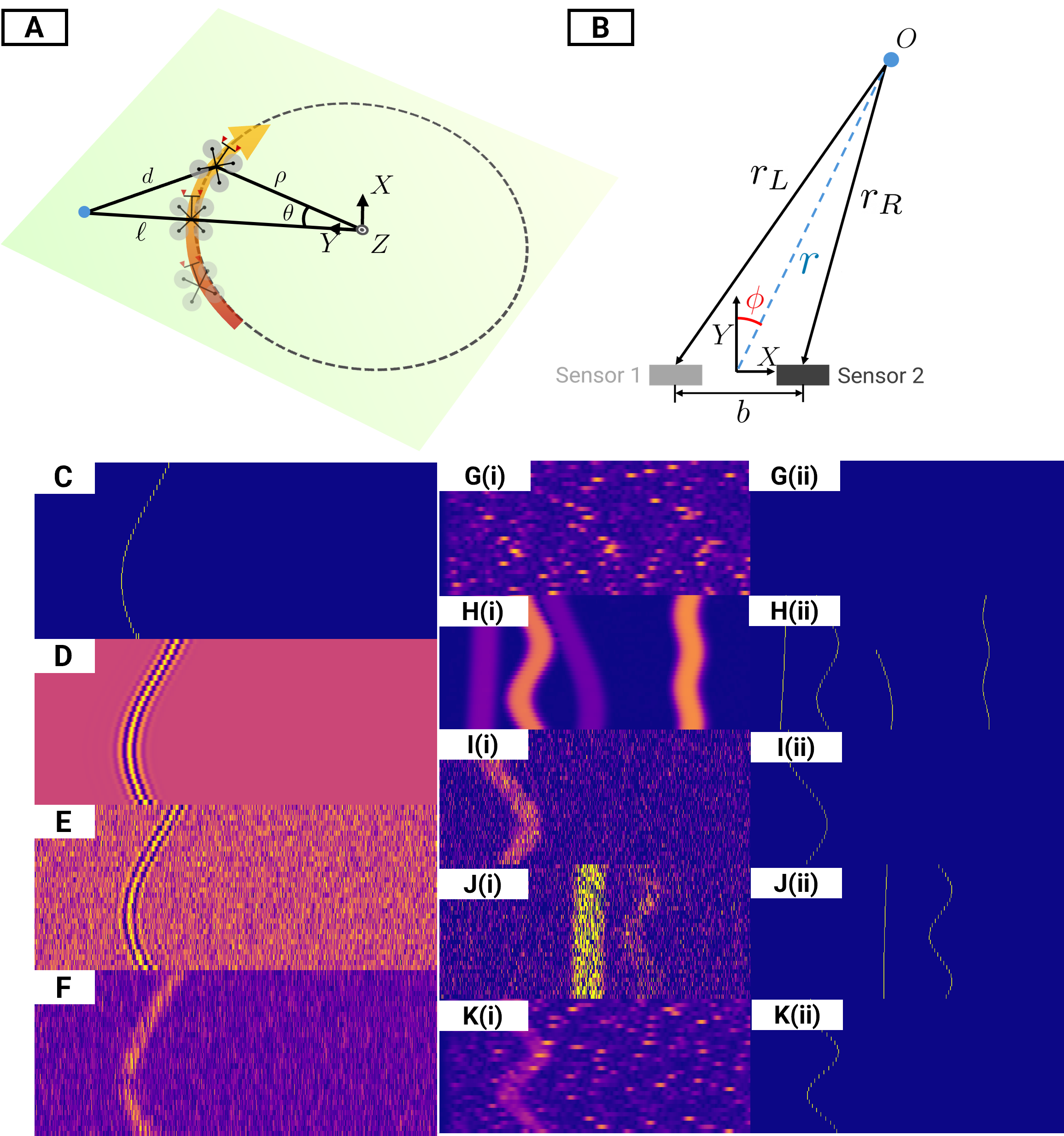}
    \caption{\textbf{Synthetic data generation pipeline for Saranga network training.} (\textbf{A}) Aerial robot following a circular arc trajectory. (\textbf{B}) Bilateration setup with dual ultrasonic sensors and obstacle $\mathcal{O}$. (\textbf{C--F}) Data synthesis process: (\textbf{C}) ideal impulse response $\mathcal{H}$, (\textbf{D}) convolution with recorded echo response $\mathcal{R}$, (\textbf{E}) noise augmentation with propeller and synthetic noise, (\textbf{F}) envelope signal $\mathcal{E}$ as network input. (\textbf{G--K}) Generated training samples $\mathcal{E}$ \textbf{(i)} with ground truth impulse responses $\mathcal{G}$ \textbf{(ii)}.}
    \label{fig:datagen_input}
\end{figure}

We modeled the robot to follow a smooth and differentiable trajectory in $\mathbb{R}^6$ in an unknown environment with multiple obstacles. The sensors can obtain echoes from a few/all obstacles, depending on the field of view of the sensor and relative sensor-obstacle positions. During a short time horizon, we approximated the trajectory using a two-dimensional circular arc \textcolor{black}{in the plane of movement} as shown in Fig.~\ref{fig:datagen_input}\textcolor{red}{A} (this is true when the changes in the direction to the plane's normal are fairly small, in most cases this corresponds to the altitude of the robot since horizontal motion is predominant). Consider the case for a single obstacle as shown in Fig.~\ref{fig:datagen_input}\textcolor{red}{B} where the sensor \textred{received} an echo for the obstacle during the entirety of the short time horizon. Without loss of generality, we defined a coordinate system at the center of the circular arc as shown in Fig.~\ref{fig:datagen_input}\textcolor{red}{A}. The distance $d$ from the ultrasound sensor on the aerial robot to the point obstacle is given by,

\begin{align}
    d=\sqrt{\left(\rho \cos\theta - (\rho+\ell)\right)^2 + \rho^2 \sin^2\theta}\\
    d=\sqrt{\rho^2+(\rho+\ell)^2-2(\rho+\ell)\rho\cos\theta}
    \label{eq:traj1}
\end{align}

\textcolor{black}{where $\rho$ represents the radius of the arc and $\ell$ is the distance from the arc to the obstacle. It should be noted that $\ell$ is collinear with $\rho$.} Using Taylor series expansion on the distance $d(\theta)$ about $\theta=0$,
\begin{align}
    d=\sqrt{\rho^2 + (\rho+\ell)^2 - 2(\rho+\ell)\rho} + \frac{\rho(\rho+\ell)(1 - \cos\theta)}{\left(\sqrt{\rho^2 + (\rho+\ell)^2-2(\rho+\ell)\rho}\right)} + \cdots
\end{align}

We neglect higher-order terms, assuming small $\theta$, 
\begin{equation}
    d \approx \rho \frac{1+(1+\ell / \rho)^2 - (1+\ell / \rho)(\cos\theta +1)}{\sqrt{(1 + \ell / \rho)^2 - \ell / \rho}}
    \label{eq:circ_eqn}
\end{equation}

If $\theta$ evolved linearly over measurement time $t$, Eq.~\ref{eq:circ_eqn} would resemble a cosine wave with a phase offset. Hence the distance/range to an $i^{\text{th}}$ obstacle in an unknown environment can be modeled as,
\begin{equation}
    d_i(t) = a_i + b_i \cos(\omega_i t+ \phi_i)
    \label{eq:trajGen}
\end{equation}

$\text{where} \
a_i \sim \mathcal{U}(a_{min}, a_{max}), \ 
b_i \sim \mathcal{U}(b_{min}, b_{max}), \
\omega_i \sim \mathcal{U}(\omega_{min}, \omega_{max}), \ 
\phi_i \sim \mathcal{U}(0, 2\pi) $.
$\mathcal{U}$ denotes uniform distribution.

Next, we modeled each ideal echo signal (trajectory in history $t$) as a cosine wave. The ideal echo signal at every time $t$ is given by the \textcolor{black}{time of flight} value of $\tfrac{2d_i(t)}{c_s}$ under the assumption that the robot speed $\ll$ speed of sound $c_s$, which is $343~\mathrm{m\,s^{-1}}$. Since our sensor operates in a discrete domain with a finite sampling rate $f$, the ideal echo signal waveform is a Dirac delta spike at the index $\left\lfloor\tfrac{2d_i(t)}{fc_s}\right\rfloor$. The normalized impulse response $\mathcal{H}$ at $t^{th}$ measurement cycle time is given by the sum of all spikes from multiple obstacles.

\begin{align}
\mathcal{H}_t = \sum_{\forall i}\delta\left(\left\lfloor{\frac{2d_i(t)}{fc_s} }\right\rfloor\right)
\label{eq:rendered_impulse_response}
\end{align}

Where $\delta(x)$ represents a Dirac delta spike that has a value of 1 at $x$ and 0 elsewhere, and $\lfloor\cdot \rfloor$ represents the floor function.


$\mathcal{H}\in \mathbb{R}_+^{M\times N}$ that represents the impulse responses for $M$ history steps and $N$ sampling measurements during each measurement cycle time (see Fig.~\ref{fig:datagen_input}\textcolor{red}{C}). 

The ideal response from the Eqs.~\ref{eq:traj1} to \ref{eq:rendered_impulse_response} is modeling point obstacles with perfect reflection characteristics for impulse input and neglecting higher-order effects that are hard to model. Training our network on these ideal obstacle responses \textred{did} not generalize to the real world since the actual sensor data \textred{looked} very different (See Fig.~\ref{fig:classical_compare}\textred{A}). To this end, we \textcolor{black}{captured} echo responses from \textcolor{black}{real world} objects, which we call $\mathcal{R}$ and \textred{convolved} with the ideal impulse function to obtain what would be read from the sensor (see Fig.~\ref{fig:datagen_input}). \textcolor{black}{$\mathcal{R}$ was captured by placing a PVC pipe, cardboard box, and play tunnel one meter away directly facing the sensor, one at a time. During data generation, one of the three echo responses was randomly selected with equal probability to improve Saranga's robustness to different \textcolor{black}{real world} objects. Although the impulse response of the sensor would vary based on the obstacle's azimuth angle, we \textred{ignored} this effect in our synthetic data generation. We observed that with different azimuth angles, the intensity of ultrasonic signals would drop, but the location of the peaks \textred{was} only dependent on the distance to the sensor.} \\To improve generalizability to the real world further, we also \textred{paid} particular attention to modeling noise from two main sources: propeller-induced noise, and noise due to environmental conditions.
Spinning propellers generate wideband noise with specifically large amplitudes in the ultrasound region, sometimes large enough to ``drown'' out the signals if the echoes are weak enough. Since generating propeller-induced ultrasound noise is hard to model explicitly, we experimentally \textred{collected} real ultrasound data by flying the robot with the sensors set to receive only mode \textcolor{black}{(physically motivated)}. \textcolor{black}{The captured noise induced by the propellers can be seen in Fig.~\ref{fig:datagen_input}\textcolor{red}{G}}. We found that just 2~minutes of flight data of noise \textred{was} enough to enable generalization to the real world. To model noise due to environmental conditions, local pressure wave turbulence and other higher-order effects, we \textred{drew} inspiration from bio-medical imaging \cite{Michailovich}. A multiplicative and additive Gaussian model \textred{was} used to mimic the speckle noise observed in ultrasound imaging \textcolor{black}{(as an engineering choice)}.

Since we only care about the echo signals rather than the actual carrier frequency variations, we \textred{computed} the envelope image $\mathcal{E}$ of the simulated ultrasonic signal to feed as the inputs to our network. Mathematically, the data generation process for the network input is summarized by
\begin{equation}
    \mathcal{E} = \left\|\mathbb{A}\left(\alpha \left\{ \mathcal{H}\star\mathcal{R} + \eta_{\mathrm{prop}}\right\} + (1-\alpha) \left\{  (1+\eta_m)(\mathcal{H}\star\mathcal{R}) + \eta_a\right\} \right)\right\|_2
\end{equation}

Where $\mathcal{E} \in \mathbb{R}_{+}^{M \times N}$ is the simulated echo signal matrix over history and measurement cycle time, $\mathbb{A}(x)=x+j\mathbb{H}(x)$ represents the analytic signal of $x$ with $\mathbb{H}(x)$ being the Hilbert transform, $\mathcal{H}$ is the ideal impulse response, $\mathcal{R}$ is the recorded echo response from a real object, $\star$ denotes the convolution operation applied row by row (per measurement cycle time), $\eta_{\mathrm{prop}}$ represents propeller noise, $\eta_m \sim \mathcal{N}(0, \sigma_m^2), \eta_a \sim \mathcal{N}(0, \sigma_a^2)$ is multiplicative and additive noise, respectively, and $\alpha \sim \mathcal{U}(0,1)$ is a binary parameter drawn from a uniform distribution that selects between the two noise models.

Finally, to obtain the ground truth, we \textred{convolved} the ideal impulse response $\mathcal{H}$ with the envelope of $\mathcal{R}$. This \textred{added} the echoes from neighboring obstacles while ignoring any interference artifacts. We then \textred{computed} the leading edge as 
\begin{align}
    \mathcal{G'}=\mathcal{H}\star\|\mathbb{A}(\mathcal{R})\|; \quad
    \mathcal{G} = \left[\mathcal{G'}(t, r)<\tau \right] \land \left[ \mathcal{G'}(t+1, r)\ge\tau \right]
\end{align}

\textcolor{black}{where $\left[ \mathcal{X} \right]$ is the Iverson bracket applied element-wise to matrix $\mathcal{X}$, returning 1 if the condition is True and 0 otherwise, \textred{which produced} a binary mask of the same dimensions.}

The ground truth $\mathcal{G}$ image would resemble the original impulse response when there \textred{were} no interferences from neighboring objects and it would have discontinuities when there \textred{were} higher-order interferences from multiple obstacles or multi-path reflections. 

When trained on input-output pairs of $\mathcal{E}$ and $\mathcal{G}$ respectively (see Fig.~\ref{fig:datagen_input}), we \textred{expected} Saranga to output a single sample width for each obstacle (think of this as one pixel-wide edge in an image); \textred{however}, we observed that the network output spans each obstacle over a few samples (akin to pixels in the image) as shown in Fig.~\ref{fig:datagen_input}. We \textred{handled} this in the next steps with post-processing before it \textred{was} used for computing control commands.

\subsubsection*{Network implementation and training}

\textcolor{black}{We \textred{drew} inspiration from M-mode imaging \cite{Feigenbaum2010} in medical ultrasound and \textred{utilized} a UNet-based \cite{ronneberger_unet2015} encoder-decoder network with four encoder and four decoder layers.} \textcolor{black}{The network takes echo images $\mathcal{E}$ as input and predicts denoised impulse responses $\tilde{\mathcal{G}}$ as output.} The network was trained in \texttt{FLOAT32} and quantized using Dynamic Range Quantization\cite{dynamic_quantization} to \texttt{UINT8} using TensorFlow$^\text{TM}$ 2.17.0 and the TF Lite Converter. The model was then compiled for acceleration on the Google Coral$^\text{TM}$ Mini using the Edge TPU Compiler\cite{edge_compiler}. We used a learning rate of $2 \times 10^{-4}$ and the \textsc{Adam} optimizer. The mini batch size was set to $128$, and the network was trained for 100~epochs using a \textcolor{black}{Mean Squared Error (MSE) loss: $\mathcal{L} = \frac{1}{N}\sum_{i=1}^{N} \|\tilde{\mathcal{G}}_i - \mathcal{G}_i\|^2$, where $N$ is the total number of elements across the batch}. \textcolor{black}{The network was trained on 40000 image pairs}. The quantized, Edge TPU-compiled model is of size 0.5~MB, with a parameter count of 1.24~M. The network is fully compatible with the Google Coral Edge TPU\texttrademark{} and \textred{took} $14.57$~ms per inference.

\begin{equation}
    \mathcal{L} = \frac{1}{N}\sum_{i=1}^{N} \|\tilde{\mathcal{G}}_i - \mathcal{G}_i\|^2
\end{equation}

\subsection*{Noise reduction}
Despite best efforts from the Saranga network, it \textred{was} \textcolor{black}{unable} to detect obstacles beyond 0.5~m distance consistently since the reflected echo signal strength \textred{was} low. This \textred{happened} mainly when the relative sensor-obstacle orientation \textred{was} large and/or the obstacle surface \textred{was} highly non-linear. In such cases, the weak echo signal \textred{got} easily overpowered by the high amount of propeller-induced ultrasound noise from vortex shedding  \cite{kurtz1970review} and diffraction\cite{PIRINCHIEVA1991183}. Note that the noise level can be reduced by requiring lower hover thrust or using larger propellers (Fig.~\ref{fig:physical_noise_reduction_ablation}). However, the hover thrust and propeller size are largely dictated by operational requirements and are not driven by sensor characteristics.

To this end, we \textred{drew} inspiration from bats\cite{suga_2005}, nature's expert echo-locators, whose calls can be as loud as 140~dB at a distance of 10~cm. Bats' middle-ear muscles contract approximately 3~ms after the laryngeal muscles initiate a call. We \textred{used} the same fundamental philosophy with implementational simplicity by using a physical shield between the propellers and the sensors (See Movie.~\ref{fig:banner}). This simple method helped improve the sensing range from one meter to two meters by \textred{reducing} the noise levels. An extensive analysis of propeller-induced noise for different shield sizes is shown in Fig.~\ref{fig:physical_noise_reduction_ablation}.

Ideally, to block all of the propeller-induced noise, an infinitely large shield would be required, which is not practically feasible. Through ablations, we \textred{identified} the size, shape and material of the shield that \textred{gave} us the sensing range for our operational constraints -- sense naturally occurring obstacles like trees at a maximum distance of 2~m.

A combination of the physical shield and a deep denoising network \textred{enabled} our robot to operate with just 10.6~mJ compute energy per inference, while being able to sense obstacles as thin as $0.025$~m at a maximum distance of $1.66$~m.


\subsection*{Obstacle localization}

As described in the Problem Definition And Contributions section, we \textred{aimed} to navigate towards a goal direction \textcolor{black}{(forward)} by avoiding obstacles (vertical obstacles by horizontal dodging) using dual low-power ultrasound chirp sensors. With noise issues \textred{mitigated} through a combination of physical noise reduction and our neural network, Saranga, we \textred{turned} our attention to the task of obstacle localization. Let the two front-facing ultrasonic sensors on our robot be called $S_L$ and $S_R$ (see Fig.~\ref{fig:datagen_input}\textcolor{red}{B}). The left sensor $S_L$ emits a short ultrasonic pulse and both sensors record the returning waveform over a small interval. Once the measurement cycle completes, the next one begins.

We \textred{stacked} the waveforms from the past $M$ measurement cycles to construct an echo image $\mathcal{E}$. Sample echo images are shown in Fig.~\ref{fig:classical_compare}, which \textred{were} fed into the Saranga network for denoising to obtain $\hat{\mathcal{G}}$. For obstacle localization, first we \textred{had} to find the correspondences between the obstacle return echoes in both the sensors $S_L$ and $S_R$. The slight difference in time between the two sensors \textred{enabled} us to localize the obstacle\cite{walter_locating_2014, nagashima_1992, Kreczmer10, THONGUN2015480}. 

We \textred{found} the non-zero indices in the predicted $\hat{\mathcal{G}}$ by a simple thresholding operation $J = \mathmybb{1}\left(\left[\hat{\mathcal{G}} > \tau_c\right]\right)$, where $\mathmybb{1}$ returns the indices of elements that are 1. Note that these correspond to the impulse responses we would obtain if the object were a ``perfect'' point reflector, essentially giving us a denoised version of the input. 
The non-zero indices of $J$ \textred{were} extracted and multiplied by $\frac{c_s}{f}$ to get a list $P$ containing the ultrasonic path length to the detected obstacles. 

We \textred{obtained} two path-length sets, one for each sensor: $P_L$ and $P_R$. To perform a correspondence search, we \textred{drew} inspiration from stereo matching in computer vision and signal processing \cite{stereo_2001} and \textred{utilized} a co-occurrence matrix obtained by taking the outer product between $P_L$ and $P_R$ sets ($P_L\otimes P_R$), which \textred{gave} us a matrix containing all possible matches between the detected obstacles from the two sensors. We then \textred{filtered} this matrix using a maximum acceptable path disparity threshold ($|p_L-p_R|\leq b_H$) to identify valid correspondences. For each valid match, we \textred{stored} the path disparities against the corresponding distance in the $P_L$ set.

As we assumed in the Saranga section, the aerial robot flew with minimal changes normal to the plane of navigation in a short time horizon while navigating (commonly the vertical axis). \textred{We analyzed} a simple scenario with one obstacle as shown in Fig.~\ref{fig:datagen_input}\textcolor{red}{B} and \textred{derived} angular position $\phi$ as a function of path disparity $\Delta p_H=p_L - p_R=(r_L+r_L) - (r_L+r_R)=(r_L-r_R)$, where $r_L$ and $r_R$ are the distance from the obstacle to the left and right sensors respectively. Here, $p_L \in P_L$ and $p_R \in P_R$ represent the matched path measurements from the corresponding sensors after filtering the outer product matrix with the acceptable path disparity matching criterion $|\Delta p_H|\leq b_H$. The distances $r_L, r_R$ are related to the obstacle's Cartesian coordinates $\mathcal{O}_x, \mathcal{O}_y$ (forward and lateral positions relative to the robot) and the baseline $b$ (see Fig.~\ref{fig:datagen_input}\textcolor{red}{B}) by, 

\begin{equation}
r_L^2 = (\mathcal{O}_y+b/2)^2 + \mathcal{O}_x^2; \quad 
r_R^2 = (\mathcal{O}_y-b/2)^2 + \mathcal{O}_x^2 \Rightarrow \mathcal{O}_y = \frac{(r_L - r_R)(r_L+r_R)}{2b_H}
\end{equation}
Assuming that the radial distances are much larger than baseline $r_L, r_R>>b_H$, the radial $r$ and azimuth angle $\phi$ of the obstacle are given by, 
\begin{equation}
    r = (r_L + r_R)/2; \quad 
    \sin\phi = \frac{\mathcal{O}_y}{r} = \frac{(r_L - r_R)}{b_H}; \quad
    \phi=\sin^{-1}\left( \frac{\Delta p_H}{b_H} \right)
    \label{theta_calc}
\end{equation}

Solution to the above equation exists only if the path disparity $|\Delta p_H|\leq b_H$. Since our approach \textred{worked} with multiple obstacles, we \textred{stored} the results (azimuth angle $\phi$ and range $r$) for each obstacle in an obstacle buffer $M_\mathcal{O}$.

Although Saranga ideally produced single sample width edges, in practice, its output often \textred{spanned} multiple samples. This \textred{resulted} in multiple candidate matches in $P_R$ for each detected edge in $P_L$. 

To handle this one-to-many correspondence problem, we \textred{computed} the azimuth angle $\phi$ using Eq.~\ref{theta_calc} for each valid match pair. Since a single obstacle in $P_L$ \textred{could} have multiple corresponding matches in $P_R$, this \textred{produced} a set of azimuth angle estimates $\{\phi_1, \phi_2, ..., \phi_n\}$ for each obstacle. We then \textred{applied} a median filter to this set of azimuth angles associated with each element in $P_L$, which \textred{provided} a robust final azimuth angle estimate by removing outliers caused by spurious matches or measurement noise.

\subsubsection*{Extension to 3D}
\textred{Although} horizontal obstacle avoidance \textred{demonstrated} the viability of our perception approach, aerial robots operate in three-dimensional environments \textred{where} both horizontal and vertical avoidance are required. To enable 3D trilateration, we \textred{augmented} our sensor configuration with an additional sensor $S_D$ positioned vertically below the reference sensor $S_L$ with baseline $b_V$.  

Extending the approach from 2D localization, we \textred{computed} a pair of path disparities from {$\{S_L, S_R\}$} and $\{S_L, S_D\}$ and \textred{applied} path disparity constraints $|\Delta p_H| \leq b_H$ and $|\Delta p_V| \leq b_V$ respectively. For each valid matching triplet $(p_L, \Delta p_H, \Delta p_V)$, we \textred{computed} the spherical coordinates:

\begin{equation}
\phi = \sin^{-1}\left(\frac{\Delta p_H}{b_H}\right); \quad
\theta = \sin^{-1}\left(\frac{\Delta p_V}{b_V}\right); \quad
r \approx p_L/2
\end{equation}

Similar to the 2D case, Saranga's output may span multiple samples, resulting in multiple candidate matches in both $P_R$ and $P_D$ for each detected edge in $P_L$.  We \textred{applied} median filtering to obtain robust final estimates $\{\hat{\phi}, \hat{\theta}\}$, effectively removing outliers from spurious matches or measurement noise. Algorithm~\ref{alg:obstacle_detection_3d} in the Supplementary Materials details the complete 3D perception process. For the 2D case, all the quantities related to the vertical sensor (color coded in \textcolor{blue}{blue}) are absent.

\subsection*{Navigation approach}
We \textred{wanted} our aerial robot to fly towards a goal direction while dodging static obstacles. In this case, the goal direction \textred{was} forward ($V_d$, 0, 0), though it could \textred{have been} derived from a global planner or navigation waypoints in a complete system - such integration is beyond the scope of this work. In the previous step, we computed the two-dimensional (optionally, 3D) location of all the detected obstacles in the plane of the robot trajectory. To simplify further, for this navigation task, we only \textred{considered} the closest detected obstacle from the $P_L$ set for dodging. We describe the navigation policy for 3D below. However, we \textred{did} not use $V_z$ (or any vertical motion-related computations) for our experiments except for the 3D one.

The robot \textred{was} commanded a velocity to dodge obstacles using a strategy inspired by the classical potential field planner\cite{kala_2024}. The controller \textred{maintained} forward progress towards the goal through the desired velocity $V_d$, while generating lateral and vertical avoidance maneuvers. 

For the closest detected obstacle with spherical coordinates $(r, \phi, \theta)$, we first \textred{converted} to Cartesian coordinates:

\begin{equation}
x = r\cos(\theta)\cos(\phi); \quad
y = r\cos(\theta)\sin(\phi); \quad
z = r\sin(\theta)
\end{equation}

\noindent where $x$ represents forward distance, $y$ lateral distance, and $z$ vertical distance to the obstacle. For obstacle avoidance, we \textred{used} the closest detected obstacle to determine forward deceleration, lateral and vertical commands ($V_x$, $V_y$, $V_z$). These \textred{were} computed as,

\begin{equation}
V_x = V_d - K_x\frac{x}{r^3}
\end{equation}
\noindent where $V_d$ is the desired forward velocity, $K_x$ is the repulsive gain for forward direction.

For lateral and vertical avoidance, we \textred{determined} the avoidance direction based on the obstacle's position in the $y$-$z$ plane. When an obstacle \textred{posed} a collision risk (distance $\|[y,z]^T\|_2 < \delta_{max}$), we \textred{computed} the avoidance direction as a unit vector pointing away from the obstacle:

\begin{equation}
u_{avoid} = -\frac{[y, z]^T}{\|[y, z]^T\|_2}
\end{equation}

This strategy naturally \textred{adapted} to the obstacle's position: obstacles with larger vertical separation $|z|$ \textred{were} avoided primarily vertically, \textred{whereas} those appearing laterally ($|z| \approx 0$) \textred{triggered} horizontal avoidance. However, obstacles directly ahead where $\|[y,z]^T\|_2 \to 0$ \textred{sometimes caused} oscillatory behavior due to measurement noise. To address this, we \textred{implemented} hysteresis based on a distance threshold $\delta$. The avoidance direction $u_{avoid}$ \textred{updated} only when $\|[y,z]^T\|_2 > \delta$; otherwise, it \textred{maintained} its previous value. The lateral and vertical velocity commands \textred{were} computed as:

\begin{equation}
[V_y, V_z]^T = \begin{cases}
[K_y, K_z]^T \cdot u_{avoid} & \text{if } \|[y,z]^T\| < \delta_{max} \\
\mathbf{0} & \text{otherwise}
\end{cases}
\end{equation}

\noindent Here, $K_y, K_z$ are the repulsive gains for the horizontal and vertical directions, respectively. The complete navigation approach is summarized in Algorithm \ref{alg:navigation_3d} in the Supplementary Materials. All the quantities in \textcolor{blue}{blue} \textred{were} only used for the 3D experiment.

\subsection*{Statistical analysis}

\begin{figure}
    \centering
    \includegraphics[width=0.91\linewidth]{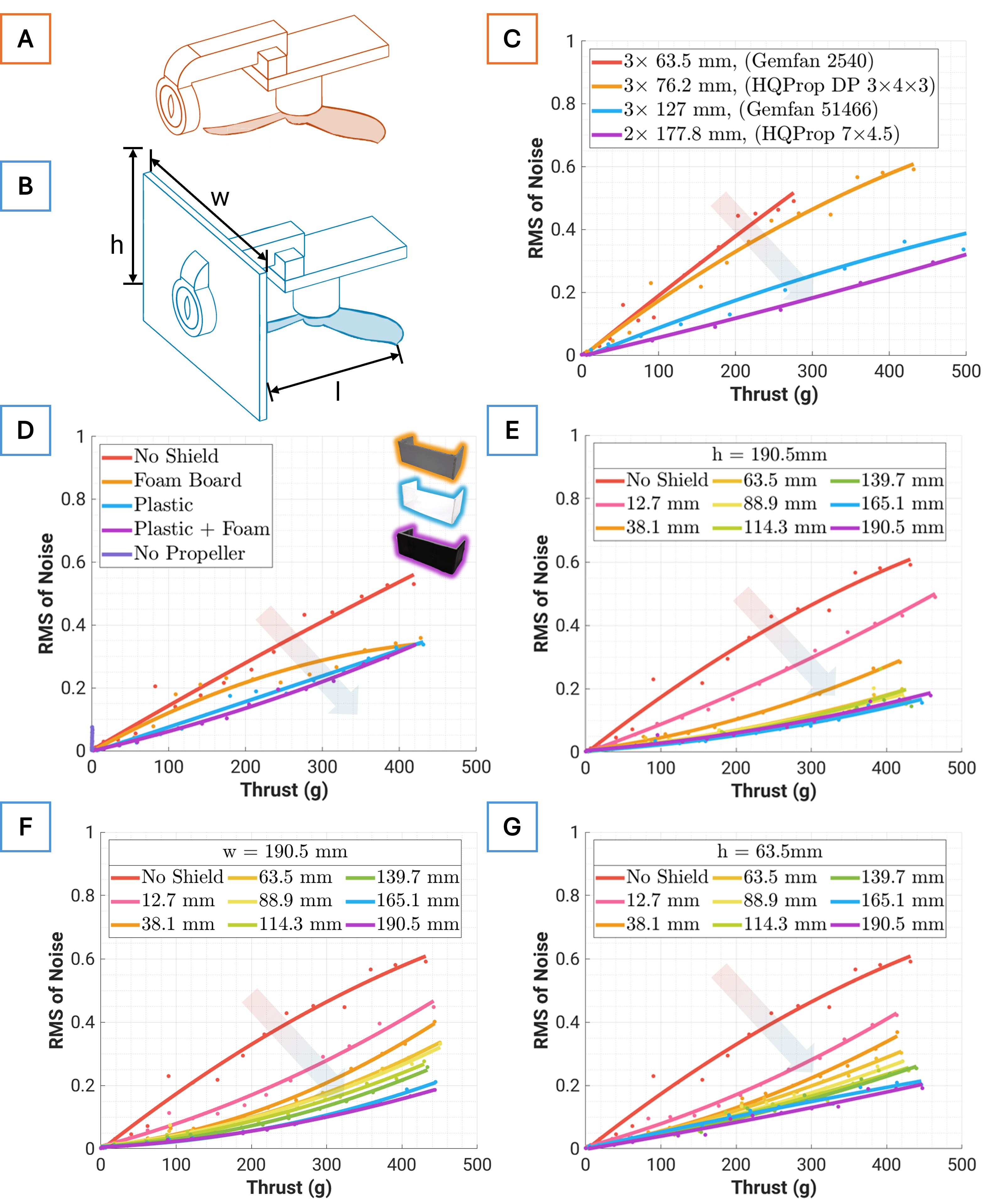}

    \caption{\textbf{Physical noise reduction ablation studies.} Noise analysis setups with (\textbf{B}) and without shielding (\textbf{A}). (\textbf{C--G}) RMS of noise measurements for increasing thrust. (\textbf{C}) Propeller-induced noise without shielding. (\textbf{D--G}) \textcolor{black}{Shield design ablation studies using 3$\times$ 76.2~mm propeller (HQProp DP 3$\times$4$\times$3").} (\textbf{D}) Shield material comparison: 4.5~mm polystyrene foam, 0.5~mm plastic, and 0.5~mm plastic with 4~mm Aggsound foam (constant dimensions: $w=$88.9~mm, $h=$63.5~mm). (\textbf{E}) Shield width variation with fixed height and material. (\textbf{F}) Shield height variation with fixed width and material. (\textbf{G}) Width variation with \textcolor{black}{the height that balanced noise reduction and size} (63.5~mm), and constant material.}
\label{fig:physical_noise_reduction_ablation}
\end{figure}

\begin{figure}
    \centering
    \includegraphics[width=0.81\linewidth]{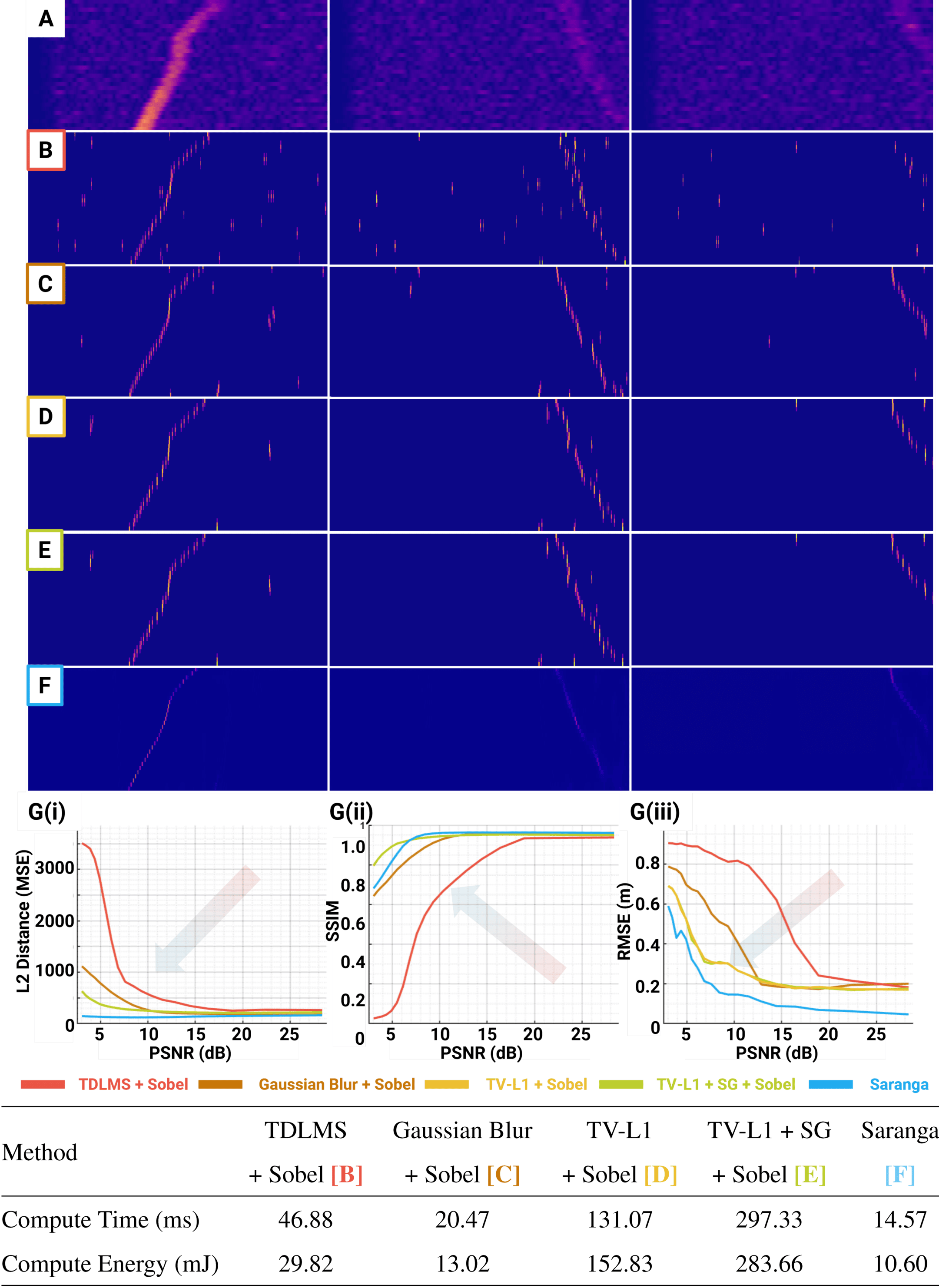}
    \caption{\textbf{Comparison of Saranga with various classical methods.} (\textbf{A}) Ultrasonic envelope signal $\mathcal{E}$ for \textcolor{black}{60~mm} $\diameter$ PVC pole. (\textbf{B--F}) Edge detection outputs: (\textbf{B}) TDLMS+Sobel\cite{hadhoud1988two}, (\textbf{C}) Gaussian Blur+Sobel\cite{davies_machine_1990}, (\textbf{D}) TV-L1+Sobel\cite{chambolle2004algorithm}, (\textbf{E}) TV-L1+SG+Sobel\cite{savitzky1964smoothing}, (\textbf{F}) Saranga. (\textbf{G}) Performance metrics: accuracy (\textbf{G(i)},\textbf{G(ii)}) and bilateration error (\textbf{G(iii)}). The table shows the compute time and energy on Google Coral Mini per network inference. Notice that Saranga output is thinner and has less noise. Further, with increasing noise, Saranga performs favorably.}
    \label{fig:classical_compare}
\end{figure}

The data plotted in Fig.~\ref{fig:physical_noise_reduction_ablation} was obtained from the experiments described in the Physical noise reduction section. All the plots were normalized with a single constant corresponding to the measurement limit (20000~AU) of the sensor, where AU denotes Arbitrary Units. The actual points obtained from the experiment are given by the dots, and the lines are quadratic fits computed using \texttt{polyfit} in MATLAB. Fig.~\ref{fig:classical_compare} experiments are described in the Comparison against denoising methods section. The line plots from the simulation study were plotted in MATLAB. The compute time was obtained by averaging the wall-clock time elapsed over 100 iterations, measured using the Google Coral's internal clock. During benchmarking, no other parts of the autonomy stack were executed, and pre-captured data were fed into the Saranga neural network and classical methodologies. To measure the compute energy per iteration, we measured the power consumption while running the neural network and subtracted the baseline power consumption measured during idle operation. This was then multiplied by the computed time per iteration. 





\begin{movie}
    \centering
    \includegraphics[width=\linewidth]{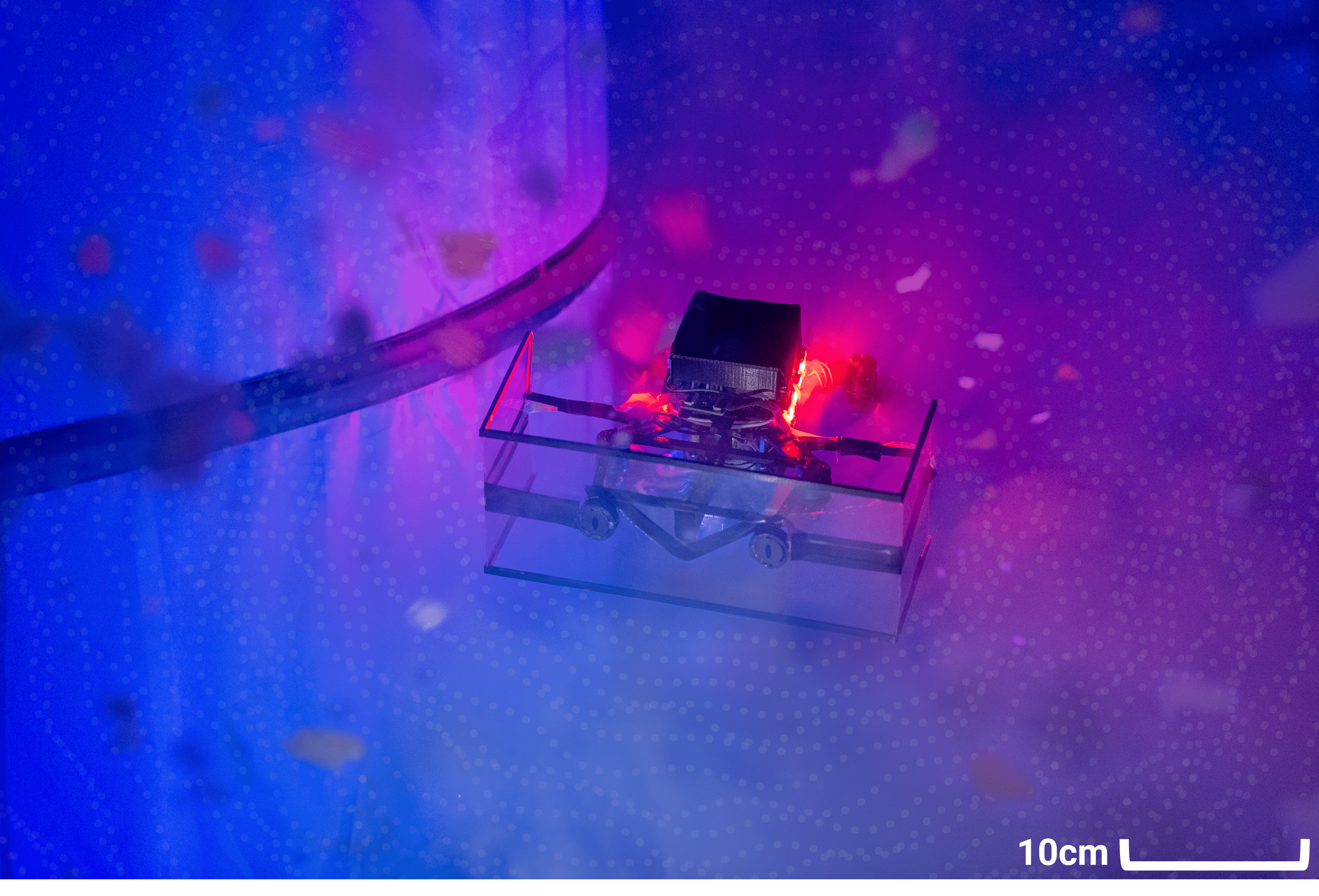}
    \caption{\textbf{Autonomous aerial robot navigation in snow, fog, and darkness using Saranga.} Our palm-sized aerial robot equipped with dual ultrasonic sensors showing obstacle avoidance in challenging visual conditions. The overlay of sound like patterns are an artistic expression to show that our work uses ultrasound signals and no vision.}

    \label{fig:banner}
\end{movie}

\begin{figure}
    \centering
    \includegraphics[width=0.84\textwidth]{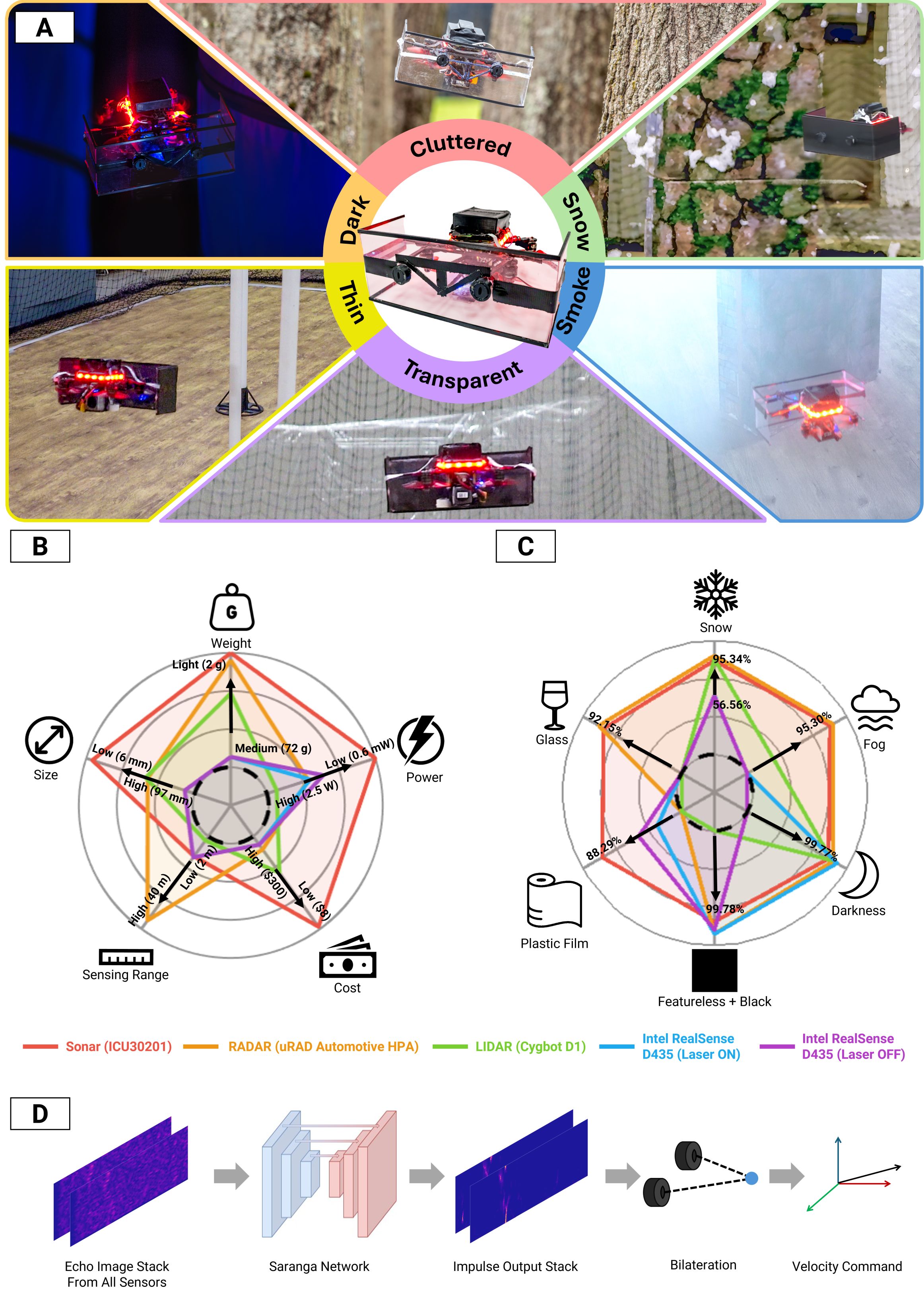}
    \caption{\textbf{System overview and sensor comparison.} (\textbf{A}) Navigation in darkness, clutter, snow, \textcolor{black}{fog}, and with transparent/thin obstacles. (\textbf{B}) \textred{Sensor comparison where lower weight, size, cost, power consumption and higher range are desirable qualities}. (\textbf{C}) Sensor performance across environmental conditions (snow, fog, darkness, glass, plastic and featureless). \textcolor{black}{(A larger area occupied by a sensor means better performance.)} (\textbf{D}) Ultrasonic processing pipeline from raw echo signals to velocity commands.}
    \label{fig:sensor_comparison_complete}
\end{figure}

\begin{figure}
    \centering
    \includegraphics[width=0.80\linewidth]{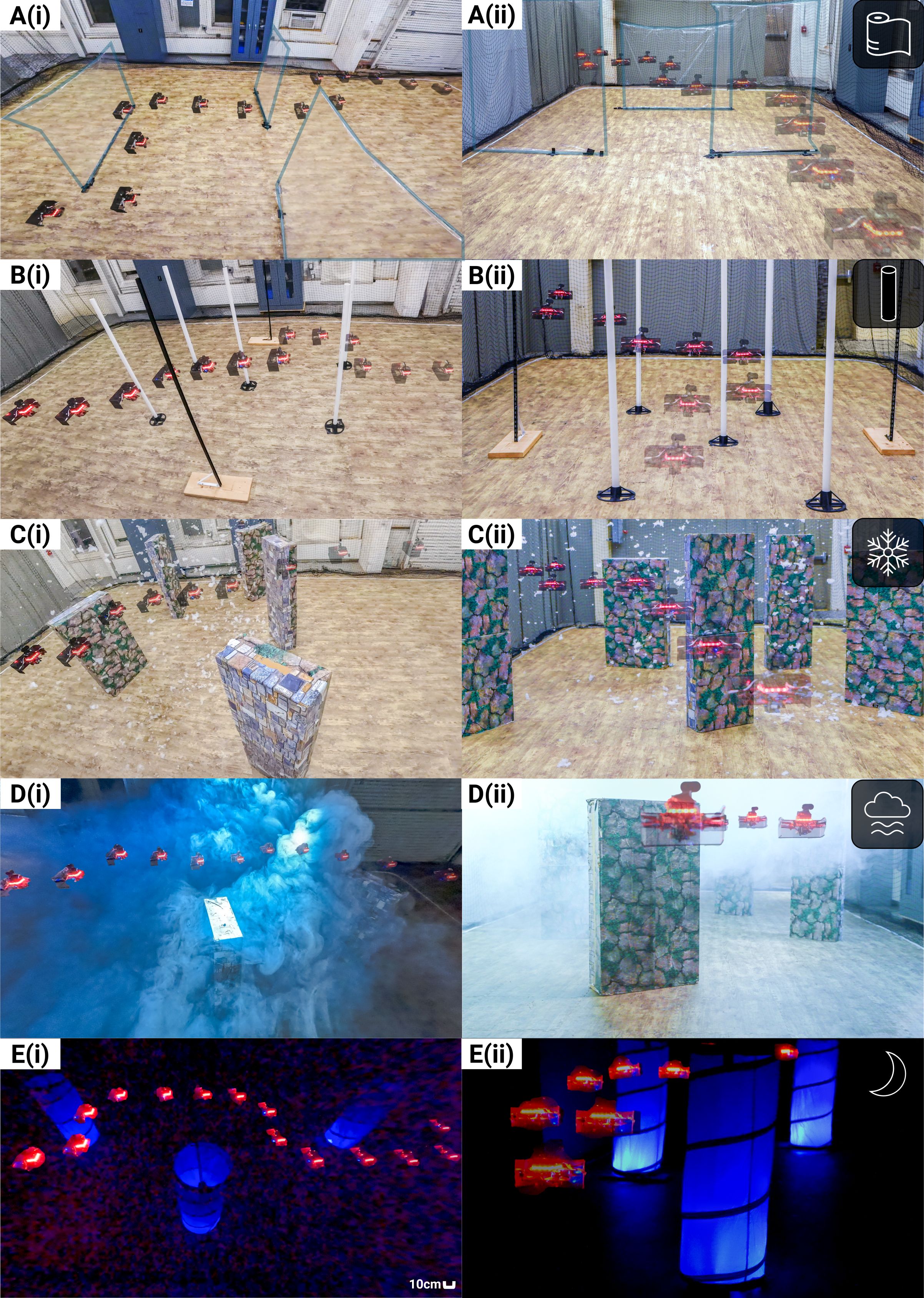}
    \caption{\textbf{Aerial robot navigation through various indoor scenes:} (\textbf{A}) Transparent obstacles \textcolor{black}{(\textcolor[RGB]{21,96,130}{Blue} highlights show the boundary of transparent film)}, (\textbf{B}) Thin objects, (\textbf{C}) Snow, (\textbf{D}) \textcolor{black}{Fog}, (\textbf{E}) Low light (dark) conditions. \textred{For each environment, \textbf{(i)} are the perspective view, \textbf{(ii)} shows the front view\textred{. For both, the opacity shows the time progression, where lower opacity is closer to the beginning of the trajectory}. The scale bar for the trajectories is shown in \textbf{E(i)}.}}
    \label{fig:indoor_nav_image}
\end{figure}

\section*{List of Supplementary Material}
\subsubsection*{This PDF file includes:}
Supplementary \textred{Methods}\\
Figures S1 to S3\\
Algorithms 1 to 2\\
Table S1

\subsubsection*{Other supplementary materials for this manuscript:}
Movies S1 to S2\\
The data plotted in Figs.~\ref{fig:sensor_comparison_complete},\ref{fig:physical_noise_reduction_ablation},\ref{fig:classical_compare} are available in the supplementary data file (Data S1).

\clearpage 

%
\bibliography{ref}
\bibliographystyle{sciencemag}

%
%
%
%
%
%


\textbf{Acknowledgments:} We thank Kushagra Srivastava for his assistance with computing camera poses using the COLMAP utility in outdoor settings. We are grateful to Deepak Singh and Hrishikesh Pawar for their support in computing the traversability scores. We acknowledge TDK InvenSense for generously providing the ICU30201 sensors used in this work. We also thank Ashwin Disa for his contributions in developing initial data collection setups. We also thank Prof. Guanrui Li for his thrust bench setup used in the experiments. \textbf{Funding:} Support by the National Science Foundation's Foundational Research in Robotics program through award number CMMI 2516439 is gratefully acknowledged. \textbf{Author contributions:} M.V. formulated the main ideas, performed formal analysis, developed the software, conducted the experiments and wrote the draft of the paper. P.B. designed the hardware, performed software development, conducted the real-world experiments and ablation studies, performed formal analysis, and reviewed and edited the manuscript. C.B. performed formal analysis, software development, created visualizations, and performed the experiments. R.J.P. provided technical guidance, resources, validation, and review and editing of the manuscript. N.J.S. conceptualized the problem, provided technical guidance, validation, funding, supervision, administration and contributed to the paper writing and analysis of the results. \textbf{Competing interests:} R.J.P. is an employee of InvenSense Inc., which produces ICU-30201, and is an inventor of several patents (e.g., US Patent 10,700,792 and US Patent 10,751,755) related to the sensor design and operation. The other authors declare that they have no competing interests. \textbf{\textred{Data, code, and materials availability:}} All (other) data needed to evaluate the conclusions in the paper are present in the paper or the Supplementary Materials. All data needed to evaluate the conclusions in the paper are present in the paper or the Supplementary Materials. \textred{The video, data, codebase of the work can be found at the project webpage \url{https://pear.wpi.edu/research/saranga.html}}. \textred{The codebase, generated datasets and trained models can be found at \url{https://doi.org/10.5061/dryad.f1vhhmh9z}.} \textred{All materials for the aerial robot and experimental setups were obtained from commercial sources.}


\newpage


\renewcommand{\thefigure}{S\arabic{figure}}
\renewcommand{\thetable}{S\arabic{table}}
\renewcommand{\theequation}{S\arabic{equation}}
\renewcommand{\thepage}{S\arabic{page}}
\setcounter{figure}{0}
\setcounter{table}{0}
\setcounter{equation}{0}
\setcounter{page}{1} 




\begin{center}
\section*{\centering \NoCaseChange{Supplementary Materials for\\ MilliWatt Ultrasound for Navigation in \\Visually Degraded Environments on Palm-Sized Aerial Robots}}

Manoj~Velmurugan$^{1}$, 
Phillip~Brush$^{1}$, 
Colin~Balfour$^{1}$, 
Richard~J.~Przybyla$^{2}$, 
Nitin~J.~Sanket$^{1\ast}$ \\ 
\vspace{0.5em}

\small$^{1}$Perception and Autonomous Robotics (PeAR) Group, Worcester Polytechnic Institute, MA, USA. \\
\small$^{2}$TDK InvenSense, Berkeley, CA, USA.\\
\vspace{0.5em}

\small$^\ast$Corresponding author. Email: \texttt{nitin@wpi.edu}\\
\end{center}

\subsubsection*{This PDF file includes:}
Supplementary \textred{Methods}\\
Figures S1 to S3\\
Algorithms 1 to 2\\
Table S1

\subsubsection*{Other supplementary materials for this manuscript:}
Movies S1 to \textcolor{red}{S2}\\
\textred{\noindent \textbf{Movie S1.}
This video shows the experiments conducted in Table 1: Snow, Thin Obstacles, Low-Light, Transparent Obstacles, Fog, 3D, Sparse Forest, Medium Forest, and Dense Forest.\\
\noindent \textbf{Movie S2.}
This video shows demonstrations of 3D obstacle avoidance in low-light, fog, and snow in a scene with dark obstacles and thin poles.}\\

The data plotted in Figs.~\ref{fig:sensor_comparison_complete},\ref{fig:physical_noise_reduction_ablation},\ref{fig:classical_compare} are available in the supplementary data file (Data S1).

\newpage


\begin{figure}
    \centering
    \includegraphics[width=0.5\linewidth]{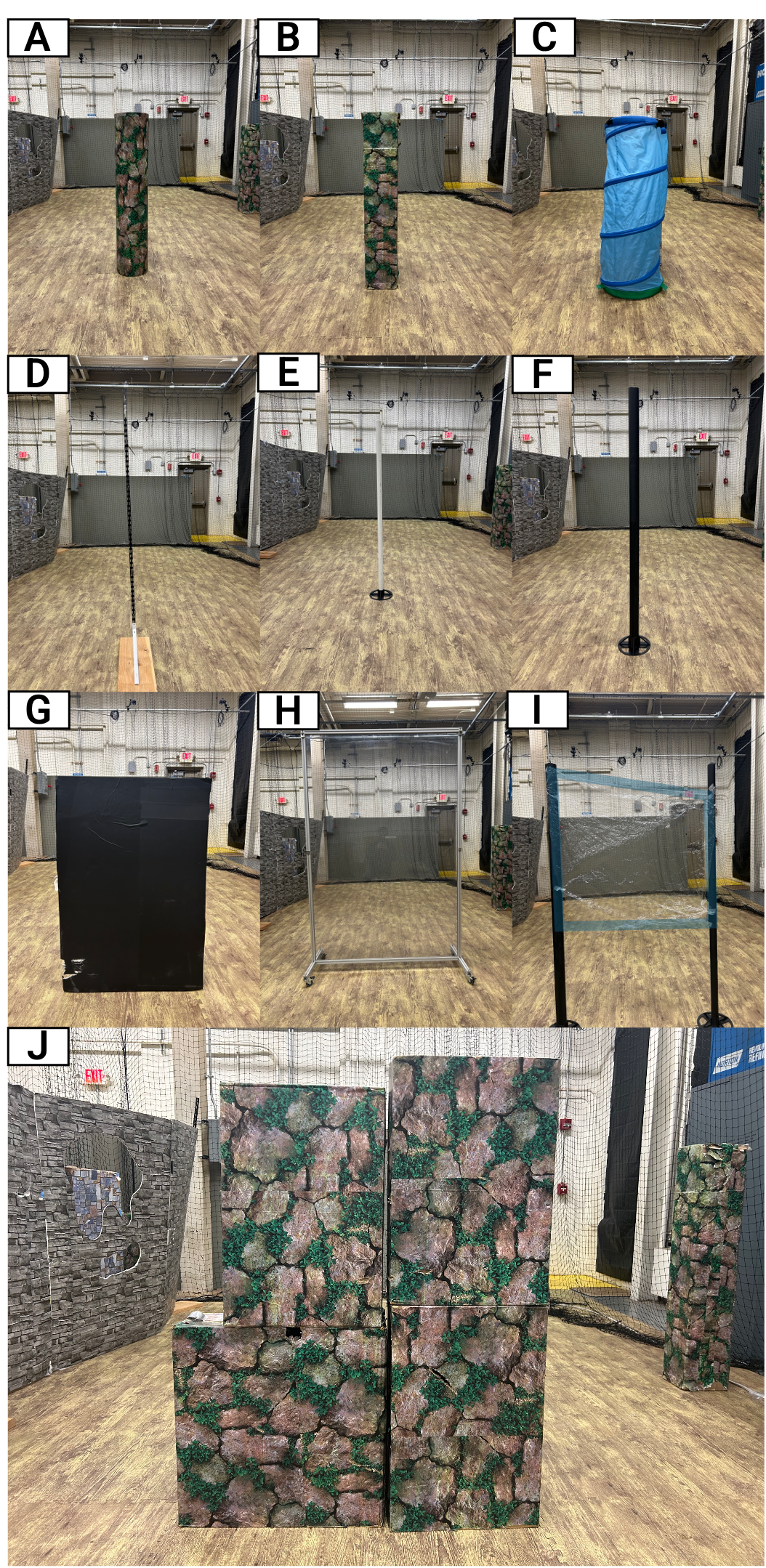}
    \caption{\textbf{Obstacles used for various experiments in indoor flight runs and comparisons with traditional robotics sensors.} \textbf{(A)} Cardboard cylinder with moss texture, \textbf{(B)} Cardboard box with moss texture, \textbf{(C)} Blue-colored translucent play tunnels, \textbf{(D)} Thin metal pole, \textbf{(E)} Thin PVC pole of 40mm$\diameter$, \textbf{(F)} Thin PVC pole of 60mm$\diameter$ with matte black texture, \textbf{(G)} Foamcore with black matte texture, \textbf{(H)} Glass wall, \textbf{(I)} Thin transparent plastic film (\textcolor[RGB]{21,96,130}{Blue} highlights show the boundary of transparent film), \textbf{(J)} Cardboard boxes with moss texture.}
    \label{fig:obstacle_images}
\end{figure}

\begin{figure}
    \centering
    \includegraphics[width=1\linewidth]{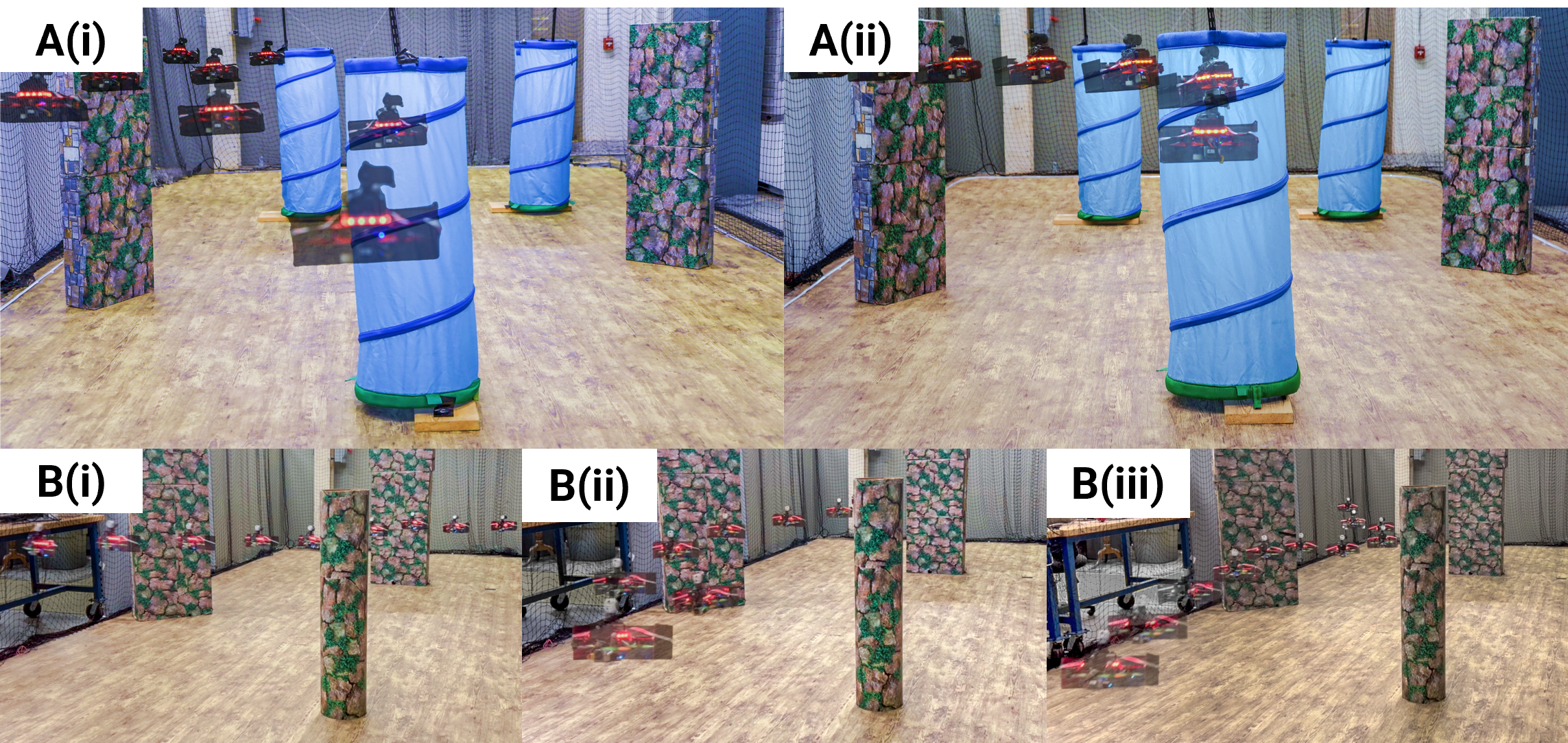}
    \caption{\textbf{Speed test and state-of-the-art comparison experiments.} (\textbf{A}) Aerial robot traversing indoor experiment environment to compare Saranga against Batdeck. Comparison of Saranga \textbf{(A(i))} with state-of-the-art BatDeck \cite{muller_batdeck_2024} \textbf{(A(ii))} in a cluttered environment. (\textbf{B}) Testing Saranga at various speeds of 1~$\mathrm{m\,s^{-1}}$ \textbf{(B(i))}, 1.5~$\mathrm{m\,s^{-1}}$ \textbf{(B(ii))}, and 2~$\mathrm{m\,s^{-1}}$ \textbf{(B(iii))}. Opacity shows time progression.}
    \label{fig:highspeed_batdeck_comparison}
\end{figure}

\begin{figure}
    \centering
    \includegraphics[width=0.75\linewidth]{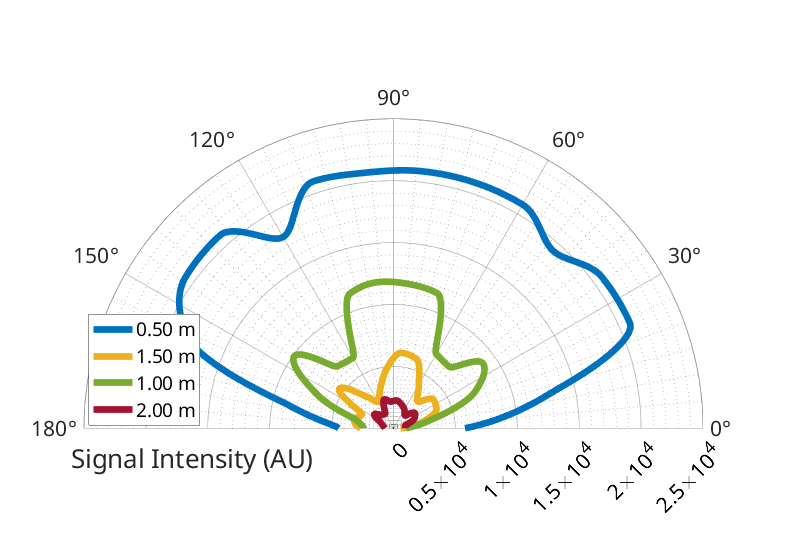}
    \caption{\textbf{Acoustic beam pattern of ultrasound sensor.} Intensity over distance and azimuth variations (beam pattern) for a thin pole of $\diameter$ 60~mm and a single ultrasound sensor.}
    \label{fig:beam_pattern_distance}
\end{figure}

\section*{Supplementary Methods}
\subsection*{Sensor comparison methodology}

In order to generate the plots for Figs. \ref{fig:sensor_comparison_complete}\textcolor{red}{B, C}, we tested various traditional robotics sensors at a range of 1m, as described in the Results section. The obstacles used are shown in Fig. \ref{fig:obstacle_images}. For snow, fog, and low-light, we used two stacked cardboard boxes with a mossy stone textured wallpaper (Fig. \ref{fig:obstacle_images}\textcolor{red}{J}). For the featureless + black obstacle (Fig. \ref{fig:obstacle_images}\textcolor{red}{G}), we used a matte black wallpaper stuck on a foamcore sheet. For the plastic film (Fig. \ref{fig:obstacle_images}\textcolor{red}{I}), we used a 0.75mil (0.02mm) thick plastic sheet attached between two PVC poles with a matte black wallpaper stuck on. Measurements on the poles were neglected in the accuracy calculations.
For glass (\ref{fig:obstacle_images}\textcolor{red}{H}), we used a large glass pane within a frame. Detections on the frame were not considered.

\subsection*{Beam pattern of ultrasonic sensor}
The beam pattern of our ultrasonic sensor can be found in Fig. \ref{fig:beam_pattern_distance}. This is the response of the sensor for various sensor distance and relative transmitted beam angle combinations. Since the sensor has a finite field of view, as the quadrotor navigates in a scene, the relative angle and distance can make it ``not see'' certain obstacles (signal strength drops below a range for reliable detection). This is one of the main modes of failure in our experiments where one or both sensors do not see the obstacles. If the obstacle appears from the sides it would be seen in the last instant which will not give enough time for dodging. For example, in the \texttt{thin obstacles} experiment shown in Fig. \ref{fig:indoor_nav_image}\textcolor{red}{B}, the scene was extremely cluttered where the robot would hit the obstacles on the side while dodging another. To obtain the beam pattern, we used a 60~mm pole mounted at different distances and azimuths as shown in \ref{fig:beam_pattern_distance}.

\begin{algorithm}[H]
\caption{\textcolor{black}{Ultrasonic Obstacle Detection and Localization in 3D}}
\label{alg:obstacle_detection_3d}
\KwIn{Echo image $\mathcal{E}_L, \mathcal{E}_R, \textcolor{blue}{\mathcal{E}_D}$ (stacked over 32 measurement cycles)}
\For{each sensor $n \in \{L, R, \textcolor{blue}{D}\}$}{
    $\hat{\mathcal{G}}_n = \text{Saranga}(\mathcal{E}_n)$ \hfill \textcolor{gray}{Network forward pass}\\
    $J_n = \mathmybb{1}\left(\left[\hat{\mathcal{G}}_n > \tau_c\right]\right)$ \hfill \textcolor{gray}{Binarization and extraction}\\
    $P_n = J_n\frac{c_s}{f}$ \hfill \textcolor{gray}{Extract path-lengths}\\
}
$M_{\mathcal{O}} = \varnothing$ \hfill \textcolor{gray}{Initialize obstacle buffer}\\
\For{$p_H \in P_L \otimes P_R$, \textcolor{blue}{$p_V \in P_L \otimes P_D$}}{
    $\Delta p_H = p_L - p_R$ \hfill \textcolor{gray}{Compute path disparity for the horizontal pair}\\
    \textcolor{blue}{$\Delta p_V = p_L - p_D$} \hfill \textcolor{gray}{Compute path disparity for the vertical pair}\\
    \If{$|\Delta p_H| \leq b_H$ and \textcolor{blue}{$|\Delta p_V| \leq b_V$}}{
        $\phi = \sin^{-1}(\Delta p_H / b_H)$ \hfill \textcolor{gray}{Compute azimuth angle}\\
        \textcolor{blue}{$\theta = \sin^{-1}(\Delta p_V / b_V)$} \hfill \textcolor{gray}{Compute elevation angle}\\
        Append $\{\phi, \textcolor{blue}{\theta}\}$ to $M_{\mathcal{O},p}$ \hfill \textcolor{gray}{Store angle estimate}\\
    }
    $\hat{M}_{\mathcal{O},p} = \text{median}(M_{\mathcal{O},p})$ \hfill \textcolor{gray}{Apply median filter}\\
}
\KwOut{$\hat{M}_{\mathcal{O}}$}
\end{algorithm}

\SetKw{KwState}{State}
\begin{algorithm}[H]
\caption{\textcolor{black}{3D Navigation Algorithm}}
\label{alg:navigation_3d}
\KwIn{Obstacle buffer $\hat{M}_{\mathcal{O}}$}
$\{r, \phi, \textcolor{blue}{\theta}\} = \arg\min_{p \in \hat{M}_{\mathcal{O}}} p$ \hfill \textcolor{gray}{Find closest obstacle}\\
$\{x,y,\textcolor{blue}{z}\}=\{r\cos(\theta)\cos(\phi),r\cos(\theta)\sin(\phi),\textcolor{blue}{r\sin(\theta)}\}$ \hfill \textcolor{gray}{Spherical to Cartesian}\\
$V_x = V_{d} - K_x\frac{x}{r^3}$ \hfill \textcolor{gray}{Apply repulsive force}\\
\If{$\|[y, \textcolor{blue}{z}]^T\|_2 > \delta$}{
    $u_{avoid} = -\frac{[y, \textcolor{blue}{z}]^T}{\|[y, \textcolor{blue}{z}]^T\|_2}$ \hfill \textcolor{gray}{Update obstacle avoidance direction}\\
}
$(V_y, \textcolor{blue}{V_z})$ = \textbf{0} \hfill \textcolor{gray}{Initialize velocity command to zero}\\
\If{$\|(y, \textcolor{blue}{z})\|_2 < \delta_{max}$}{
    $(V_y, \textcolor{blue}{V_z}) = [K_y, K_z]^T\cdot u_{avoid}$ \hfill \textcolor{gray}{Compute velocity command}\\
}
\KwOut{Velocity commands $(V_x,~V_y,~\textcolor{blue}{V_z})$}
\end{algorithm}

\begin{table}[h]
\centering
\begin{tabular}{ll}
\hline
\textbf{Component} & \textbf{Link} \\
\hline
IMONE 650W Snow Machine & \footnotesize{\url{www.amazon.com/dp/B0DF6BNSDY}} \\
IKEA DV\"ARGM{\AA}S Play Tunnel & \footnotesize{\url{www.ikea.com/us/en/p/dvaergmas-play-tunnel-blue-green-90547595}} \\
Vangoa 1500W Fog Machine & \footnotesize{\url{www.amazon.com/dp/B0CDLQV9NN}} \\
Sunolga 900W Fog Machine & \footnotesize{\url{www.amazon.com/dp/B09VGTWZX9}} \\
AggSound Sound Deadening Mat  & \footnotesize{\url{www.amazon.com/dp/B0BV2GQMSG}} \\
Yancorp Peel and Stick Wallpaper & \footnotesize{\url{www.amazon.com/dp/B07PG41YR7}} \\
\hline
\end{tabular}
\caption{Equipment used in experiments.}
\label{tab:equipment}
\end{table}

\end{document}